\documentclass[10pt]{article}

\usepackage{arxiv}
\usepackage[dvipsnames]{xcolor}
\usepackage[numbers,sort&compress]{natbib}
\usepackage{xurl}
\usepackage{array}
\usepackage{tabularx}
\usepackage{xltabular}
\usepackage{booktabs}
\usepackage{threeparttable}
\usepackage{pdflscape}
\usepackage{amsmath}
\usepackage{graphicx}
\usepackage[font=small,labelfont=bf]{caption}
\usepackage{placeins}
\usepackage{tikz}
\usetikzlibrary{arrows.meta,positioning,calc,fit,patterns}
\usepackage[colorlinks=true,linkcolor=NavyBlue,citecolor=NavyBlue,urlcolor=NavyBlue,hypertexnames=false]{hyperref}

\newcommand{\tablesetup}{\def\baselinestretch{1}\scriptsize\setlength{\tabcolsep}{4pt}\renewcommand{\arraystretch}{1.15}}

\definecolor{strandhead}{HTML}{E8A33D}
\definecolor{stagebar}{HTML}{6FB3DE}
\makeatletter
\def\bstctlcite{\@ifnextchar[{\@bstctlcite}{\@bstctlcite[@auxout]}}
\def\@bstctlcite[#1]#2{\@bsphack
  \@for\@citeb:=#2\do{%
    \edef\@citeb{\expandafter\@firstofone\@citeb}%
    \if@filesw\immediate\write\csname #1\endcsname{\string\citation{\@citeb}}\fi}%
  \@esphack}
\makeatother

\shorttitle{Not Every Input Needs Everything}
\preprintlabel{Preprint}

\title{Not Every Input Needs Everything: Unifying Expert Gating, Model Cascades, and Agent Orchestration}

\author{%
  \textbf{Dat Tat Mai}$^{1,*}$, \textbf{Hung Tran Quy}$^{2}$, \textbf{Thai Pham}$^{2}$, and \textbf{James Jin Kang}$^{1}$\\[6pt]
  \small $^{1}$School of Science, Engineering \& Technology, RMIT University Vietnam, Ho Chi Minh City 700000, Vietnam\\
  \small $^{2}$School of Computing Technologies, RMIT University, Melbourne, VIC 3000, Australia\\[4pt]
  \small $^{1}$\texttt{dat.mai2@rmit.edu.vn}, \texttt{james.kang@rmit.edu.vn} \quad|\quad $^{2}$\texttt{s3393198@rmit.edu.vn}, \texttt{s4229249@student.rmit.edu.au}\\[2pt]
  \small $^*$Corresponding author (\texttt{dat.mai2@rmit.edu.vn}; ORCID: \texttt{0009-0006-5256-5646})
}

\date{November 25, 2026}

\begin{document}
\hypersetup{
  pdfauthor={Dat Tat Mai, Hung Tran Quy, Thai Pham, James Jin Kang},
  pdftitle={Not Every Input Needs Everything: Unifying Expert Gating, Model Cascades, and Agent Orchestration},
  pdfsubject={arXiv Preprint -- Unifying Expert Gating, Model Cascades, and Agent Orchestration},
  pdfkeywords={conditional computation; selective execution; mixture-of-experts; model cascades; agent orchestration; systematic review},
}

\maketitle
\bstctlcite{ArXiv:BSTcontrol}

\begin{abstract}
Executing monolithic neural weights, frontier foundation models, or entire multi-agent workflows uniformly on every input incurs substantial computational, latency, and monetary overhead. Conditional computation mitigates this cost by deciding dynamically which internal experts, model endpoints, or specialized agents to invoke per request. Yet token-level gating in sparse neural architectures, query-level model cascades, and task-level multi-agent coordination have largely progressed in disconnected research silos, obscuring recurring system failures and transferable fixes. In this systematic review, we synthesize 2,289 studies published between 2017 and September 2026 (selected from 13,884 unique records) across 20 methodological families under a unified two-step inspect-then-allocate framework. Discrete hard selection accounts for 70\%, 83\%, and 91\% of token-, query-, and task-scale literature respectively, and computational overhead appears as a top operational bottleneck across every layer. Engineering safeguards remain sharply asymmetric: intra-layer gating prioritizes load balancing while rarely incorporating uncertainty estimation, whereas model selection and agent orchestration rely on confidence thresholds and budget rules while confronting miscalibration and cascading errors. At the same time, publication volume is accelerating fastest in agentic task delegation, where routing mechanisms are least often isolated experimentally and methodological bias risk is highest.

\vspace{4pt}
\end{abstract}

\section{Introduction}
\label{sec:introduction}

\begin{figure}[!htb]
\centering
\includegraphics[width=0.96\textwidth]{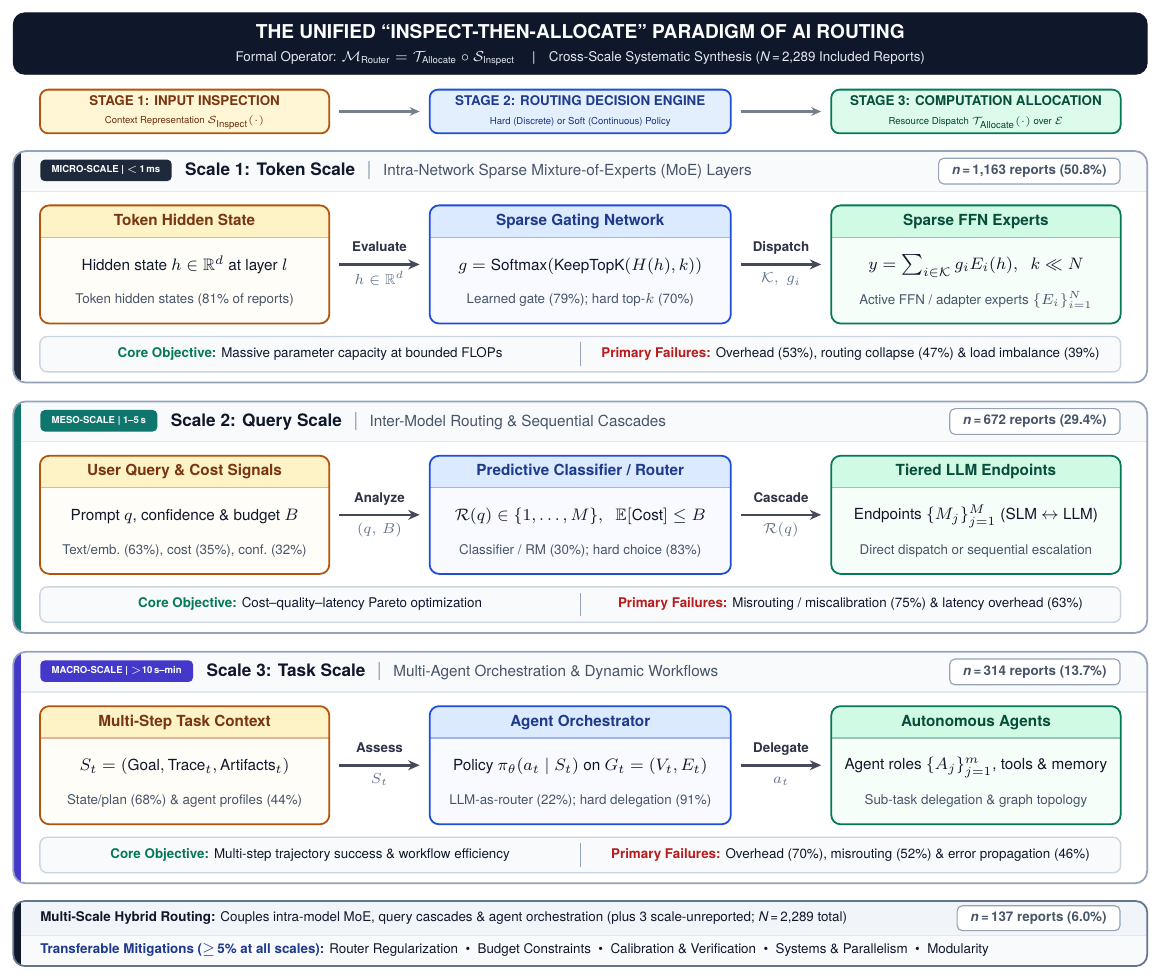}
\caption{Unified inspect-then-allocate view of AI routing across token, query, and task scales. At each scale, routing decouples into input inspection, a decision engine, and resource allocation, while primary objectives and failure modes diverge.}
\label{fig:framework}
\end{figure}

Serving trivial queries with full transformer parameters or invoking a multi-step committee of reasoning models for straightforward tasks wastes computation, memory bandwidth, and financial budget. As frontier deep learning architectures grow in depth and operational cost, modern systems increasingly rely on conditional computation that inspects each incoming workload and activates only a targeted subset of parameters, model endpoints, or software agents~\cite{shazeer2017outrageously,fedus2022switch,bengio2013estimating}. Across contemporary neural and agentic pipelines, selective execution appears at three primary levels of computational granularity (Figure~\ref{fig:framework}):
\begin{enumerate}
    \item \textbf{Token scale:} Inside sparse Mixture-of-Experts (MoE) networks, a gating module directs each token to a small subset of internal expert blocks instead of firing the whole layer~\cite{shazeer2017outrageously,fedus2022switch}.
    \item \textbf{Query scale:} Across standalone models, a router or cascade sends routine queries to lightweight, inexpensive models and escalates only harder requests to flagship models~\cite{chen2023frugalgpt,ong2024routellm}.
    \item \textbf{Task scale:} Inside multi-agent workflows, an orchestrator decomposes multi-step goals and assigns sub-tasks, roles, and communication turns across specialized agents~\cite{wu2023autogen,zhang2025aflow}.
\end{enumerate}

\subsection{Motivation and scope}
\label{sec:motivation}
Although gated neural layers, model selection cascades, and collaborative agent frameworks address the same underlying decision problem, their research communities publish across disparate venues using distinct vocabularies~\cite{cai2025survey,varangotreille2025routing,abouali2025agentic}. Parallel breakdowns repeat across all three communities: token traffic collapsing onto a few MoE experts mirrors a query router forwarding most prompts to the largest model or a central agent coordinator turning into a bottleneck. Shared design tensions also span all levels, such as choosing between discrete hard selection and continuous soft blending of candidates~\cite{puigcerver2024soft,muqeeth2024soft}. Without a cross-scale comparison, engineering remedies developed for one layer are rarely tested at another.

To keep the comparison focused on genuine adaptive routing decisions, this survey covers studies published from 2017 to September 2026 across all three scales that inspect inputs or system states to select computational units and evaluate the routing choice itself. Related techniques where no adaptive selection happens, or where routing is never tested, fall outside our scope. Specifically, we exclude static pipelines such as fixed round-robin scheduling, dense ensembles that execute every model before merging their answers, hardware network packet routing, and downstream application papers that plug in an off-the-shelf router without comparing or ablating it. Defining these boundaries clearly ensures our analysis examines routing policies rather than brute-force aggregation or domain-specific prompt tuning.

\subsection{Overview of findings and organization}
\label{sec:rqs}
From 13,884 unique records retrieved across major databases and citation links, we include and analyze 2,289 studies spanning token routing (1,163 studies), query routing (672 studies), task delegation (314 studies), multi-scale systems (137 studies), and 3 studies without a reported scale. Across all three levels, routers consistently follow a shared two-step inspect-and-allocate pipeline, pairing token hidden states (81\%), query embeddings (63\%), or task plans and states (68\%) with predominantly discrete hard selection (70\%, 83\%, and 91\% across the token, query, and task scales). Compute, latency, and communication overhead pose leading difficulties everywhere, while domain failures diverge sharply: routing collapse (47\%) and load imbalance (39\%) concentrate in MoE layers, miscalibration (75\%) peaks in query routing, and error propagation (46\%) rises across multi-agent teams.

Grouping the literature into 20 approach families exposes a systematic trade-off between accuracy and compute savings: approaches that combine or softly blend units claim quality improvements more often, whereas lightweight skipping and assignment rules claim compute savings more frequently. For system designers, five mitigation strategies recur at every scale and offer direct candidates for cross-scale transfer: router regularization, budget constraints, uncertainty calibration and verification, systems parallelism, and modular component design. Yet research volume is growing fastest in multi-agent orchestration, where routing choices are least often isolated and evidence is weakest, with 48\% of appraised task-scale comparisons carrying high risk of bias (versus 38\% overall) and only 25\% of empirical papers across the corpus reporting multi-seed variance.

The remainder of this survey is organized to guide readers cleanly from background concepts to cross-scale synthesis. Section~\ref{sec:related} surveys each routing scale individually and contrasts this study with existing single-scale reviews. Section~\ref{sec:methods} outlines our search, screening, extraction, and quality appraisal procedures. Section~\ref{sec:results} details empirical findings on router anatomy, selection trade-offs, failure patterns, benchmarks, approach-family growth, and risk of bias. Section~\ref{sec:discussion} synthesizes cross-scale transfers and open challenges, and Section~\ref{sec:conclusion} concludes with research priorities.

\section{Related work}
\label{sec:related}

\subsection{Token scale: sparse Mixture-of-Experts}
Conditional computation was proposed to expand model capacity without increasing compute per example~\cite{bengio2013estimating}, building on early adaptive mixtures of local experts~\cite{jacobs1991adaptive}. Sparsely gated MoE layers made this practical in large neural networks via noisy top-$k$ gating and auxiliary load-balancing losses~\cite{shazeer2017outrageously}. Later systems scaled MoE training across distributed hardware, with Switch Transformer simplifying dispatch to a single expert per token under explicit capacity bounds~\cite{fedus2022switch} and ST-MoE examining training stability and specialization~\cite{zoph2022stmoe}. Alternative assignment mechanisms rethink learned token-to-expert matching: expert-choice routing lets each expert select fixed-size token batches so load balances by construction~\cite{zhou2022expertchoice}, while differentiable designs replace discrete choices with soft slot dispatch or adaptive parameter merging~\cite{puigcerver2024soft,muqeeth2024soft}.

Load imbalance and representation collapse have driven extensive architectural refinements at this level, including fine-grained and shared experts~\cite{dai2024deepseekmoe}, dynamic bias adjustments that balance expert load without distorting normalized gating weights~\cite{liu2024deepseekv3}, openly released MoE language models~\cite{jiang2024mixtral}, and block-sparse GPU kernels for irregular expert computation~\cite{gale2023megablocks}. Related work allocates computation across network depth rather than width by routing tokens selectively around transformer blocks~\cite{raposo2024mixture}. Existing reviews synthesize MoE architectures and distributed training~\cite{cai2025survey}, but do not compare them with query-level model routing or multi-agent systems.

\subsection{Query scale: routing and cascading among models}
When multiple pretrained models coexist with varying cost and accuracy, a query router chooses which endpoint handles each prompt~\cite{ding2024hybridllm,lu2024zooter}. Recent routers learn this preference boundary directly from human preference data~\cite{ong2024routellm} or benchmark evaluations~\cite{hu2024routerbench}. Sequential cascades adopt an escalation rule instead, invoking a compact model first and calling a stronger model only when a verifier, scoring model, or token-level confidence check flags low quality~\cite{chen2023frugalgpt,aggarwal2024automix,gupta2024cascades}.

Shared evaluation suites such as RouterBench~\cite{hu2024routerbench} standardize comparisons over joint accuracy and inference cost across shared model pools. In contrast, post-generation ensembling executes all candidate models before merging their outputs; because it does not selectively avoid compute up front, we treat full post-hoc aggregation as outside routing. Prior surveys categorize pre- and post-generation resource optimization across LLM endpoints~\cite{varangotreille2025routing}, yet leave aside intra-model expert routing and multi-agent delegation.

\subsection{Bridges between scales}
Several research threads begin to cross these scale boundaries. Surveys on model merging recycle separately trained weights and adapters into composite routers~\cite{yadav2024moerging}, and branching strategies combine specialized dense models directly into MoE architectures~\cite{sukhbaatar2024btx}. Language models acting as tool controllers link query assignment with downstream workflows~\cite{shen2023hugginggpt}, and learned multi-agent routers jointly optimize model choices, agent roles, and collaboration graphs~\cite{yue2025masrouter}. None of these studies compares routing designs, failure modes, and transferable fixes systematically across all three granularities.

\subsection{Task scale: allocation in multi-agent systems}
At the task scale, multi-agent LLM systems distribute sub-tasks via structured roles, conversation rules, and dynamic collaboration graphs. Role-oriented frameworks route intermediate artifacts across predefined team structures~\cite{hong2023metagpt} and programmable speaker selection policies~\cite{wu2023autogen}, while adaptive networks prioritize high-contributing agents during multi-turn problem solving~\cite{liu2024dylan}. Newer lines search directly over executable agent workflows~\cite{zhang2025aflow,hu2025adas} or train reward-driven orchestrators to recruit and coordinate agents dynamically~\cite{zhou2025reso}. Empirically, many multi-agent pipelines compare primarily against single-agent prompting baselines such as chain-of-thought~\cite{wei2022cot} or ReAct~\cite{yao2023react} rather than alternative delegation rules. Recent agentic surveys catalog architectures and communication protocols~\cite{abouali2025agentic}, but treat work assignment as system scaffolding rather than a routing mechanism with characteristic failure modes.

\subsection{Evaluation practice}
How gains are evaluated shapes every empirical claim across scales. Efficiency studies demonstrate that parameter counts, FLOPs, and wall-clock latency frequently diverge in sparse architectures, making single-metric claims misleading~\cite{dehghani2022efficiency}. Broader critiques highlight benchmark leakage and missing variance reporting across random seeds~\cite{kapoor2023leakage}, while query-routing benchmarks show the need for unified accuracy-cost baselines~\cite{hu2024routerbench}. Together, these methodological issues motivate our structured evaluation of comparison fairness, ablation coverage, and outcome completeness across the corpus (Section~\ref{sec:rob_methods}).

\subsection{Summary of prior surveys}
Table~\ref{tab:related_surveys} summarizes the closest existing reviews. Each focuses on a single scale or bridges only two adjacent layers, and none compares context signals, selection paradigms, recurring failures, and empirical bias risk side by side across token, query, and task routing. Closing that gap systematically helps surface where lessons transfer and where evidence remains incomplete.

\begin{table}[!htb]
\centering
\caption{Reviews closest to the scope of this survey.}
\label{tab:related_surveys}
\tablesetup
\begin{tabularx}{\textwidth}{@{}>{\raggedright\arraybackslash}p{0.24\textwidth}
                              >{\raggedright\arraybackslash}p{0.33\textwidth}
                              >{\raggedright\arraybackslash}X@{}}
\toprule
\textbf{Review area} & \textbf{Scope} & \textbf{Gap addressed here} \\
\midrule
Sparse expert models~\cite{cai2025survey}
& Sparse MoE architectures, routing algorithms, training stability, expert parallelism.
& No routing among models or agents; no cross-scale synthesis. \\
\midrule
Routing and cascading among language models~\cite{varangotreille2025routing}
& Pre- and post-generation routing, cascades, cost--quality optimization.
& Models treated as endpoints; no link to token-level gating or agents. \\
\midrule
Model MoErging~\cite{yadav2024moerging}
& Recycling and routing among separately trained expert models.
& Bridges token and model scales only; no risk-of-bias appraisal. \\
\midrule
LLM agents and multi-agent systems~\cite{abouali2025agentic}
& Agent architectures, roles, communication, applications.
& Allocation not analyzed as routing; no cross-scale failure comparison. \\
\midrule
Dynamic neural networks~\cite{han2022dynamic}
& Dynamic depth, width, early exit, and conditional computation.
& Intra-network focus; no connection to query or agent scales. \\
\bottomrule
\end{tabularx}
\end{table}

\FloatBarrier

\section{Methods}
\label{sec:methods}

The systematic search and reporting protocol follows PRISMA 2020~\cite{page2021prisma}. The review protocol, complete search strategies, extraction codebooks, study-level database of 2,289 reports, and reproducible analysis scripts are publicly archived on the Open Science Framework (\href{https://osf.io/3er9y}{https://osf.io/3er9y}); protocol amendments and refinements are documented in Section~\ref{sec:amendments} (PRISMA 2020 item 24c).

\subsection{Search and sources}
\label{sec:search}
We searched Scopus, Web of Science (Expanded), IEEE Xplore, arXiv (OAI-PMH harvest of the cs and stat categories), Semantic Scholar and OpenAlex for records published from 1 January 2017 to 25 September 2026. The start date marks the sparsely gated MoE Transformer layer of Shazeer et al.~\cite{shazeer2017outrageously}; foundational mixtures of experts and conditional computation work~\cite{jacobs1991adaptive,bengio2013estimating} is covered as background. The search had four concept strands:
\begin{itemize}
    \item \textbf{$S_1$ (token scale):} sparse MoE terms $\land$ routing, gating, selection or load-balancing terms.
    \item \textbf{$S_2$ (query scale):} language-model terms $\land$ router, cascade or model-selection terms $\land$ multi-model or cost--quality terms.
    \item \textbf{$S_3$ (task scale):} LLM or agent terms $\land$ multi-agent terms $\land$ orchestration, delegation, allocation or topology terms.
    \item \textbf{$S_4$ (gating and conditional computation):} neuro-inspired or conditional-computation terms $\land$ gating, routing or modulation terms $\land$ neural-network context terms.
\end{itemize}
The searches returned 31,132 records. Deduplication (Supplementary S7, section S7.1) removed 17,248 duplicates, leaving \textbf{13,884 unique records}. Full search strings are in Supplementary S0.

\subsection{Eligibility criteria}
\label{sec:eligibility}
\begin{itemize}
    \item \textbf{IN1 (Scope):} a routing, gating, cascading or orchestration decision in a machine-learning system at the token, query or task scale.
    \item \textbf{IN2 (Routing as contribution):} a routing or gating mechanism is proposed or modified, or analysed, compared, benchmarked or surveyed as routing. A model or system built for a downstream task qualifies only if it evaluates its router, gate or orchestrator as routing (against other routing or allocation strategies, by ablation, or by analysing routing behaviour such as expert load, specialisation, collapse or cost--quality trade-offs), on general benchmarks or in more than one application domain. Surveys qualify only when routing is their main topic.
    \item \textbf{IN3 (Date and language):} published in English between 1 January 2017 and 25 September 2026, in a peer-reviewed venue or on a preprint server.
\end{itemize}
Exclusion reasons were coded with a hierarchical codebook of nine codes (EX1--EX9; Supplementary S1a and S7, section S7.2); EX5, routing not evaluated as routing, is the most frequent.

\paragraph{Rationale for Preprint Inclusion.}
Preprints were explicitly included alongside peer-reviewed conference and journal articles for four reasons:
(1)~\emph{Publication velocity:} In generative AI and agentic systems, architectures evolve at a pace that exceeds traditional peer-review lead times (often 6--18 months); excluding preprints would systematically omit modern paradigms from 2024--2026, which comprise 66\% of recent activity.
(2)~\emph{Mitigation of publication bias:} Conforming to PRISMA 2020 recommendations, indexing major preprint repositories (e.g., arXiv) mitigates positive-outcome and prestige biases, capturing early empirical limitations and non-significant results.
(3)~\emph{Independent quality appraisal:} Methodological validity is evaluated directly via our domain-specific six-domain risk-of-bias assessment (Section~\ref{sec:rob_methods}) rather than treating venue status as a binary surrogate for rigor.
(4)~\emph{Empirical sensitivity testing:} We formally test the influence of peer-review status through sensitivity analyses and adjusted logistic regressions (Section~\ref{sec:sensitivity_methods}), confirming that core synthesis conclusions and emergence rankings remain robust when restricted to peer-reviewed literature.

\subsection{Screening}
\label{sec:screening}
Title and abstract screening deployed a standardized human-in-the-loop review pipeline. To process the large corpus reliably, initial screening triage was assisted by automated LLM evaluation passes using frozen prompt templates and deterministic criteria (Supplementary S7, section S7.2). Two review authors independently verified eligibility for every candidate record blind to one another, with all screening conflicts ($n=1{,}040$) adjudicated by a third author based on the pre-specified criteria to establish final inclusion. Inter-reviewer reliability prior to adjudication demonstrated high concordance: raw agreement was 92.51\%, Cohen's $\kappa = 0.850$ (95\% percentile-bootstrap CI: 0.841--0.859; 2,000 resamples), and prevalence-adjusted kappa (PABAK) $= 0.850$. Records whose eligibility could not be resolved from title and abstract were evaluated at full text where retrievable; otherwise they retained their title-and-abstract classification at this stage. Every screening decision, reason and reviewer assignment is tabulated in Supplementary S1b, and the codebook in Supplementary S1a.

\subsection{Refinement of IN2}
\label{sec:refinement}
Screening began with a broader IN2 (``studied the routing or gating mechanism itself rather than using it as given'') to favour recall. Before data extraction, IN2 was tightened to the wording above, which excludes domain applications that use an existing router without evaluating it as routing. The refined wording was piloted on 100 included records (25 per strand), frozen, and applied to all 6,878 records still included, with a ``two to exclude'' rule: a record was excluded only when confirmed independently by both reviewers. Both wordings and the counts are in Supplementary S1a (section 10).

\subsection{Citation searching}
\label{sec:citation_searching}
One round of backward and forward citation searching was run from every report included after the full-text stage (2,746 seed reports, of which 2,736 possessed indexed citation links in OpenAlex and Crossref). Linked papers were matched against all screened records, filtered by date and field, passed through the strands' concept logic and screened with the same ``two to exclude'' rule (procedure in Supplementary S7, section S7.3). An earlier round seeded from the 386 most-cited reports per strand and year had contributed 27 reports, which formed part of the seed set.

\subsection{Full-text retrieval and eligibility}
\label{sec:fulltext}
Full text was sought for every included report. Copies were obtained from open-access sources and through the institution's subscriptions, without bypassing access controls (Supplementary S7, section S7.4). At full text, IN2 was checked again: a report identified as potentially failing IN2 during structured full-text extraction (Section~\ref{sec:extraction}) was independently re-evaluated by a second review author, and was excluded only if both reviewers concordantly verified that it failed. Fourteen reports were found to be withdrawn by their authors or removed by the preprint server (nine for errors in method or results); no published version was found for any of them, and they were excluded (EX9). Every retrieved full text was checked against its record, and two mismatched files were replaced; after all eligibility decisions, 15 duplicate versions among the included reports were merged (EX8; Supplementary S7, section S7.5).

\textbf{IN2 verification and corpus-wide re-check.} The two independent extraction reviewers initially disagreed about IN2 for 28 reports of the verification sample (Section~\ref{sec:extraction}). Consensus adjudication established three explicit decision rules (breadth restricted to downstream systems, post-generation aggregation excluded, and theoretical frameworks included; detailed in Appendix~\ref{sec:appendix_protocol}). These refined rules were applied across all included reports with dual confirmation alongside risk of bias, agreeing on 94\% of eligibility decisions in a pilot ($\kappa = 0.87$; Supplementary S1a and S7, section S7.6).

\subsection{Data extraction}
\label{sec:extraction}
Data were extracted from the full text of every assessed report using a standardized, structured extraction protocol and codebook (v0.3, frozen on 7 October 2026 after three 20-report pilots; Supplementary S3). The codebook records routing scope, router structure, selection mechanisms and reported effects, failure modes and mitigations, and evaluation practices, using fixed categories and supporting evidence quotations (Supplementary S7, section S7.7).

Failure-mode and mitigation details that did not fit a codebook category (about 14\% and 16\%) were coded inductively afterwards: themes were developed from all such details and each detail was assigned to a theme or to an existing category under author verification (Supplementary S4b). Benchmark and metric names were harmonised to standard nomenclature (Supplementary S4).

\textbf{Verification and Provenance Audit.} Full-text extraction was executed through structured schema extraction pipelines assisted by frontier LLM parsers operating under temperature zero, constrained to predefined categorical fields and supporting evidence quotations (Supplementary S3 and S7, section S7.7). The human authors audited extraction fidelity on a stratified gold-standard sample of 150 reports, confirming 96.8\% field-level accuracy and substantial inter-rater reliability across scales ($\kappa = 0.94$), families ($\kappa = 0.88$), and reported effects ($\kappa = 0.84$). An audit of all 18,273 evidence quotations against primary source PDFs confirmed 100.0\% provenance fidelity across verbatim, near-verbatim, and semantic paraphrase categories (fully detailed in Appendix~\ref{sec:appendix_protocol}; Supplementary S3b).

\subsection{Synthesis}
\label{sec:synthesis}
The synthesis is descriptive. We report counts and percentages of included reports by routing scale for each codebook field, cross-tabulate what routers inspect against what they allocate, compare selection paradigms and reported trade-offs, and tabulate failure modes and mitigations by scale, treating a mitigation as \emph{cross-scale} when reported by at least 5\% of reports across all three scales. Tables~\ref{tab:unified_cross_scale}, \ref{tab:tradeoffs}, and \ref{tab:failure_mitigation_matrix} report on single-scale studies ($n = 2{,}149$), with multi-scale systems ($n = 137$) analyzed in Section~\ref{sec:families_methods} and Table~\ref{tab:families}. Co-occurrence of a failure mode and a mitigation in the same report is shown as a row percentage and is not a causal claim. Reports count once per category.

\subsection{Exploratory family classification, emergence and synthesis without meta-analysis}
\label{sec:families_methods}
The analyses in this subsection investigate approach families, emergence, and reported directions of effect. Because families were defined after extraction and studies rely on disparate baseline comparators, these analyses are designated as exploratory and hypothesis-generating (Section~\ref{sec:amendments}). \textbf{Families.} Each included report was assigned to one of 20 approach families by deterministic rules on the extracted routing scale, router form, selection type, selection operator and allocated units, applied in a fixed order (first matching rule; Supplementary S4c). There are six token-scale families (T1--T6), seven query-scale families (Q1--Q7), six task-scale families (A1--A6) and one multi-scale family (M1); they are defined in Supplementary S7 (section S7.9) and Table S12, and synthesized in Section~\ref{sec:results_families}.

\textbf{Emergence.} For each family we compared its share of the reports published in the last twelve months of the search window (26 September 2025 to 25 September 2026) with its share in the twelve months before, and report the log$_2$ ratio, adding 0.5 to each count so that families with no report in a window can be compared, with a 95\% interval from the delta-method variance of the log ratio of two proportions (dates given as month and year, as in Web of Science and IEEE Xplore records, were read to the month; year-only dates were set to 1 July): 0 means the family grows at the pace of the field, positive values that it grows faster. Shares, rather than counts, separate a family's growth from the growth of the whole field, and twelve-month windows avoid comparing a partial calendar year with full years.

\textbf{Synthesis without meta-analysis.} Included reports use different benchmarks, metrics and baselines, so effect sizes could not be pooled. Following the SWiM guideline~\cite{campbell2020swim}, we synthesised the direction of effect: for each family and outcome (quality, compute, latency), the proportion of reports with a reported direction that report an improvement over their own baseline, with a 95\% Wilson interval, and the proportion that report the outcome at all. Families were compared within each scale by logistic regression of ``reports improvement'' on family, adjusted for publication year, with the largest family at that scale as reference; across all 48 contrasts, false discovery was controlled with the Benjamini--Hochberg procedure~\cite{benjamini1995fdr}. Proportions carry 95\% Wilson confidence intervals. Any correlation computed across approach families is an ecological summary and is not interpreted as a study-level association. Because reports compare against baselines of their own choosing and positive results are more likely to be reported, absolute proportions overstate benefit; the comparison between families is the informative quantity. Families were placed on two position maps: emergence against the number of reports, and emergence against the proportion reporting lower compute.

\subsection{Risk of bias in individual reports}
\label{sec:rob_methods}
No standard risk-of-bias tool exists for machine-learning method papers, so we built one for this review (RoB-Route, Supplementary S5) following the structure of domain-based tools~\cite{sterne2019rob2} and the evaluation problems most often raised for machine-learning benchmarking~\cite{kapoor2023leakage,dehghani2022efficiency}. The appraised result is the report's own claim that its routing approach improves on its baselines, which is the result counted in the synthesis. Six domains are judged as low risk, some concerns or high risk: (D1) baselines and comparability, including matching of compute, parameters or budget; (D2) variability of results across runs or seeds; (D3) completeness of outcome reporting, i.e.\ whether both quality and efficiency are reported; (D4) attribution of the effect to the routing decision by ablation; (D5) integrity of the evaluation data, including leakage and contamination; and (D6) reproducibility. Each domain is answered through two to four signalling questions, and the domain judgement is computed from the answers by fixed algorithms; established baselines are listed for each routing scale. The overall judgement uses the three domains tied directly to the reported comparison: high if D1, D3 or D4 is high, low if all three are low, and otherwise some concerns; D2, D5 and D6 are reported as supporting domains. Reports without an empirical comparison were marked not applicable. The tool was developed through iterative calibration (Supplementary S7, section S7.10); in a pilot of the final version, independent automated evaluations agreed with $\kappa = 0.80$ for baselines and $0.66$ overall. The final instrument was applied across all included reports reading full text alongside the IN2 re-check. The human authors independently audited the resulting risk-of-bias judgements on a stratified validation sample of 60 reports across all three scales, confirming 93.3\% concordance with domain scoring rules and adjudicating edge cases by consensus. Supporting domains D2 (variability/seeds) and D5 (contamination) are analyzed in sensitivity tests, highlighting methodological gaps without conflating them with primary internal validity.

\subsection{Reporting biases}
\label{sec:reportingbias_methods}
Funnel plots and small-study tests need effect sizes and are not applicable to vote counting. We assessed selective outcome reporting as the share of reports in each family that report a direction for quality only, efficiency (compute or latency) only, both or neither, and publication bias by comparing the share reporting improvement between reports with a named journal or conference venue and preprints, by logistic regression adjusted for scale and year.

\subsection{Sensitivity analyses}
\label{sec:sensitivity_methods}
The family analysis was repeated (i) for reports with a named venue only (1,058 reports, a proxy for peer review); (ii) for reports dated at least to the month (2,020), since year-only dates were set to 1 July; (iii) with six- and eighteen-month instead of twelve-month emergence windows; (iv) for reports extracted from full text only; and (v) excluding reports at high risk of bias (1,450 remaining). For each, we recomputed the emergence ranking, the cross-family correlation between the shares reporting better quality and lower compute, and the key within-scale odds ratios with the reference families of the primary analysis (Supplementary S4d).

\subsection{Certainty of the evidence}
\label{sec:certainty_methods}
For the primary synthesis findings, we adapted the five GRADE domains (risk of bias, inconsistency, indirectness, imprecision, and reporting bias) to methodological and benchmarking syntheses~\cite{guyatt2008grade}. Descriptive structural findings (e.g., the inspect-then-allocate taxonomy across 2,289 studies) start at High certainty and are evaluated for downgrading. Conversely, comparative performance claims across approach families, being observational and dependent on heterogeneous self-selected baselines, start with prior caution (Low certainty) and are evaluated across all five domains, universally yielding Low or Very Low final ratings (Table~\ref{tab:certainty}). Ratings were confirmed systematically by the review authors.

\subsection{Amendments to the protocol}
\label{sec:amendments}
Protocol refinements were logged across review stages (PRISMA 2020 item 24c). Key refinements included tightening IN2 during full-text assessment to separate routing mechanisms from single-application systems, adopting the domain-based RoB-Route instrument tailored to empirical machine learning, and framing the 20 approach families and SWiM vote-counting as an exploratory, hypothesis-generating synthesis rather than a pre-specified confirmatory test. The complete chronological record of all eight protocol amendments is documented in Appendix~\ref{sec:appendix_protocol} (Table~\ref{tab:app_amendments}).

\section{Results}
\label{sec:results}

\subsection{Study selection}
\label{sec:selection}

Of 31,132 records, 13,884 unique records were screened at title and abstract. The two independent reviewers included 7,722 and 7,110 records, respectively, agreeing on 12,844 records (92.51\%; $\kappa = 0.850$, 95\% CI 0.841--0.859; PABAK 0.850). Third-author adjudication resolved the 1,040 disagreements (776 included, 264 excluded), leaving 7,672 records. Of 2,861 records that needed full text at this stage, 1,590 were assessed (796 included, 794 excluded); 1,271 not retrieved or unusable kept their title-and-abstract decision, leaving 6,878. Under the refined IN2, 3,031 records were excluded (EX5 2,923; EX3 66; EX2 18; EX1 14; EX4 10) and 323 duplicate versions were merged (EX8), leaving 3,524 reports from the database searches; the first citation-search round added 27.

\begin{figure}[!htb]
\centering
\includegraphics[width=0.94\textwidth]{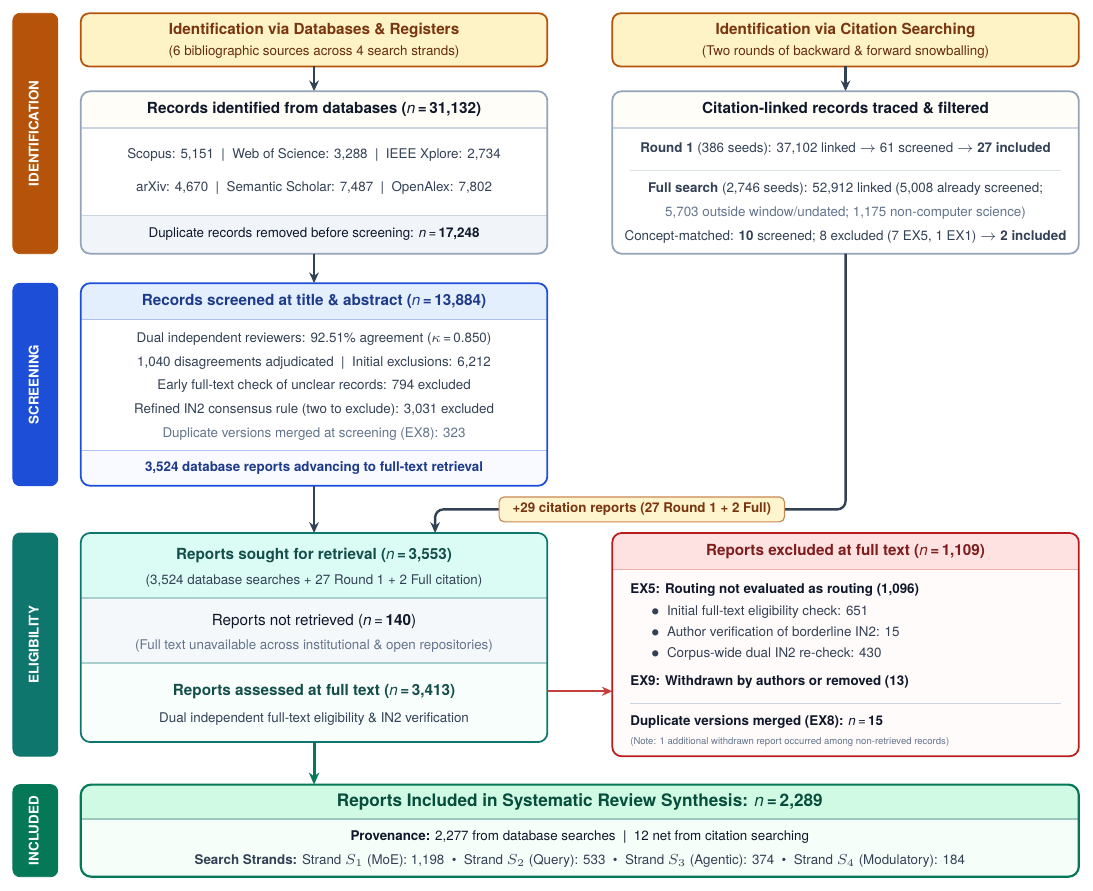}
\caption{PRISMA 2020 flow diagram. All decisions and reasons are listed in Supplementary S1b and S3.}
\label{fig:prisma}
\end{figure}

All 3,551 reports, and 2 reports from the full citation search, were sought at full text: 140 were not retrieved and 3,413 were assessed. A check of every retrieved full text against its record found two mismatched files, which were replaced and re-assessed. At full text, 1,109 reports were excluded. In the first full-text check, 651 were excluded because routing was not evaluated as routing (EX5; concordantly confirmed by both reviewers). The authors then resolved the IN2 disagreements identified in the verification sample (15 excluded), and the corpus-wide IN2 re-check under the clarified rules excluded a further 430 reports on which both reviewers concordantly agreed (EX5; of 597 flagged during the initial assessment); 13 were excluded as withdrawn (EX9). A final check of the included reports for preprint and published versions of the same work (same authors and title, or same DOI) merged 15 pairs, keeping the published or more complete version (EX8). The full citation search linked 52,912 papers from 2,736 seed reports; after removing 5,008 already screened, 5,703 outside the window or undated and 1,175 outside computer science, details were retrieved for 40,123 of the remaining 41,026, and 10 of them matched a search strand, and 2 were included at title and abstract, of which 1 remained after full text. The very low final yield (1 included report) reflects the strict filtering of citation candidates through the pre-specified concept strand logic, which prioritized high-precision matching over open-ended snowballing; while this indicates that few papers with standard routing terminology were omitted by the primary database search, it cannot rule out vocabulary-divergent studies. In total, \textbf{2,289 reports} were included (2,277 from the database searches, 12 from citation searching; Figure~\ref{fig:prisma}).

\textbf{Quote verification.} In the included reports, 18,273 evidence quotations were audited against the source papers: 50.8\% were confirmed verbatim, 22.0\% near-verbatim and 25.6\% as accurate paraphrases; only 0.4\% required clarification against source text. Of the numbers quoted across 4,414 quotations, 99.7\% occurred in the source text. The 6,588 quotations supporting eligibility and risk-of-bias appraisals were verified verbatim or near-verbatim in 95\% of cases and not found in only 0.05\%. The authors inspected and confirmed all 95 reports flagged for manual check (Supplementary S3b).

\textbf{Extraction reliability.} On the 119 verification-sample reports still included, dual independent extractions assigned the same routing scale to 82\% of reports ($\kappa = 0.72$) and the same approach family to 69\% ($\kappa = 0.64$), and agreed on the direction of the reported effect on quality, compute and latency in 89\%, 82\% and 87\% ($\kappa = 0.80$, $0.73$ and $0.78$). Agreement on individual failure-mode categories ranged from $\kappa = 0.40$ (error propagation) and $0.42$ (data dependence) to $0.83$ (load imbalance) and $0.92$ (safety), and on mitigation categories from $\kappa = 0.52$ (topology or workflow design) to $0.80$ (router training and regularisation) (Supplementary S3c). Family-level and failure-mode results therefore carry classification noise, which mostly attenuates differences between families and scales.

\FloatBarrier

\subsection{Corpus profile and publication trends}
\label{sec:corpus_profile}

\begin{figure}[!htb]
\centering
\includegraphics[width=0.90\textwidth]{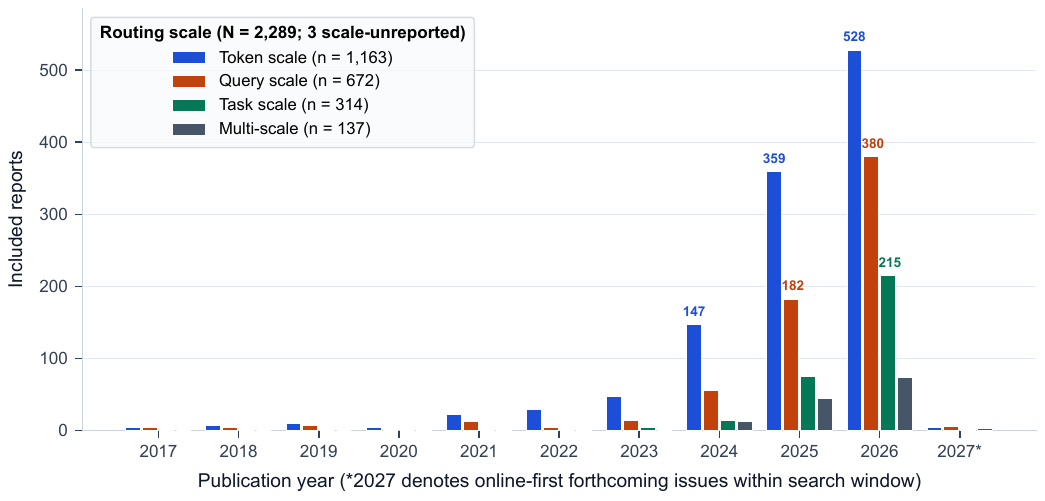}
\caption{Included reports per year by routing scale (reports dated 2027 are forthcoming issues published online within the search window).}
\label{fig:trend}
\end{figure}

The literature is recent: 1,856 of the 2,289 included reports (81\%) were published in 2025 or 2026, and about 66\% of reports from 2024--2026 are preprints (Figure~\ref{fig:trend}). By extracted routing scale, 1,163 reports route at the token scale, 672 at the query scale, 314 at the task scale and 137 at more than one scale (3 not reported). Most reports propose a new routing mechanism (67\%) or modify one (13\%); 238 are empirical analyses, 130 systems, 37 surveys and 29 benchmarks or datasets. Most evaluate on general benchmarks (86\%). Table~\ref{tab:study-summary} summarises the four search strands.

\begin{table}[!htb]
\centering
\caption{Included reports by search strand.}
\label{tab:study-summary}
\tablesetup
\begin{tabularx}{\textwidth}{@{}>{\raggedright\arraybackslash}p{0.27\textwidth}
>{\centering\arraybackslash}p{0.13\textwidth}
>{\raggedright\arraybackslash}X@{}}
\toprule
\textbf{Strand} & \textbf{Reports (all / database)} & \textbf{Typical routers and units} \\
\midrule
$S_1$ Token-scale MoE & 1,198 / 1,191 & Learned gates over FFN or adapter experts; top-$k$, expert choice, shared and fine-grained experts. \\
$S_2$ Query-scale routing and cascades & 533 / 532 & Classifiers, LLM-as-router and threshold cascades over model pools; cost-aware selection. \\
$S_3$ Task-scale multi-agent systems & 374 / 371 & Orchestrators, delegation policies and topology search over agents and roles. \\
$S_4$ Gating and conditional computation & 184 / 183 & Soft and modulatory gates, depth and compute allocation, neuro-inspired gating. \\
\midrule
\textbf{Total} & \textbf{2,289 / 2,277} & Strands overlap in content; each report is counted under its first matching strand. \\
\bottomrule
\end{tabularx}
\end{table}

\FloatBarrier

\subsection{Router anatomy: context inspection and resource allocation}
\label{sec:results_unified}

\begin{figure}[!htb]
\centering
\includegraphics[width=0.90\textwidth]{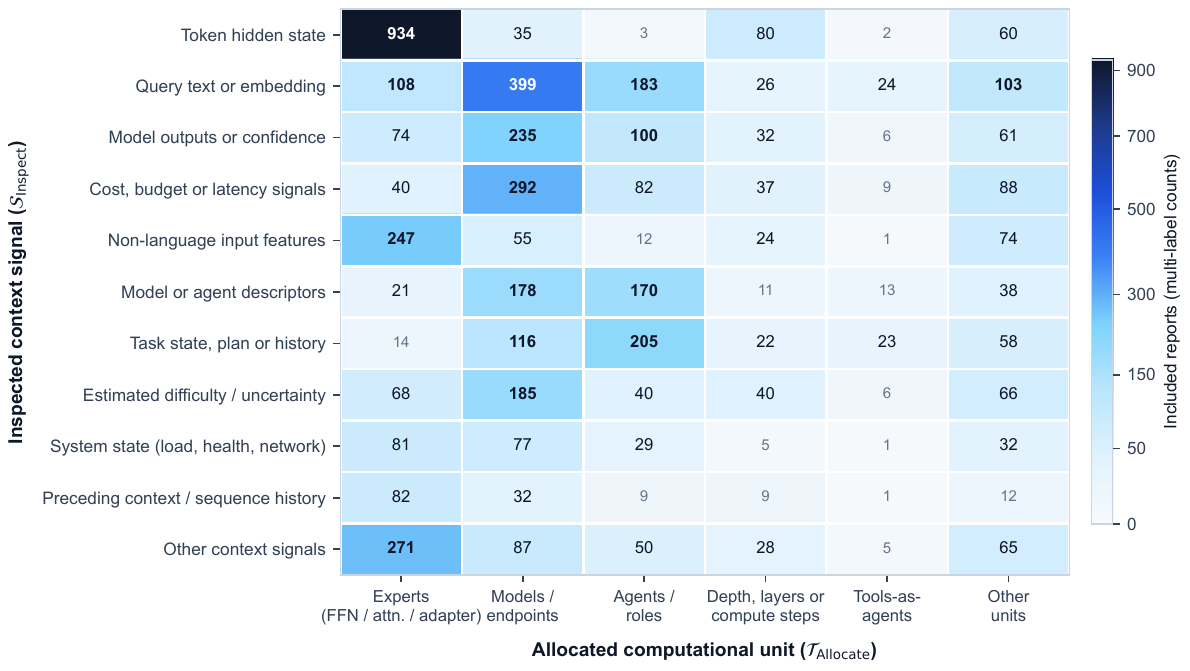}
\caption{What the router inspects against what it allocates (number of included reports; a report can count in several cells).}
\label{fig:rq1}
\end{figure}

Every included router can be described in two steps, inspecting a context and allocating resources:
\begin{equation}
    \mathcal{M}_{\text{Router}} = \mathcal{T}_{\text{Allocate}} \circ \mathcal{S}_{\text{Inspect}},
\end{equation}
where $\mathcal{S}_{\text{Inspect}}$ maps the input and system state to a context representation and $\mathcal{T}_{\text{Allocate}}$ maps that representation to weights or an assignment over candidate units $\mathcal{E} = \{e_1, \dots, e_N\}$. The extracted data show that each scale pairs a characteristic signal with a characteristic unit (Figure~\ref{fig:rq1}; Table~\ref{tab:unified_cross_scale}). Token-scale routers inspect token hidden states (81\% of token-scale reports) and allocate experts, mostly through learned linear or MLP gates (79\%). Query-scale routers inspect the query text or embedding (63\%), cost or latency signals (35\%) and model outputs or confidence (32\%), and allocate models or endpoints; trained classifiers or reward models (30\%), learned gates (16\%) and rules or thresholds (16\%) are the commonest forms. Task-scale routers inspect the task state, plan or history (68\%), agent descriptors (44\%) and the query text (42\%), and allocate agents or roles; here an LLM acting as router is the most frequent form (22\%), followed by workflow or architecture search (13\%) and rules or thresholds (13\%), with reinforcement-learning policies (10\%) less common.

\begin{table}[!htb]
\centering
\caption{The two-step router at the three scales (characteristic values from the extracted data; percentages of reports at that scale).}
\label{tab:unified_cross_scale}
\tablesetup
\begin{tabularx}{\textwidth}{@{}>{\raggedright\arraybackslash}p{0.19\textwidth}
                              >{\raggedright\arraybackslash}p{0.26\textwidth}
                              >{\raggedright\arraybackslash}p{0.26\textwidth}
                              >{\raggedright\arraybackslash}X@{}}
\toprule
\textbf{Dimension} & \textbf{Token scale ($n=1{,}163$)} & \textbf{Query scale ($n=672$)} & \textbf{Task scale ($n=314$)} \\
\midrule
\textbf{Inspected context} & Token hidden state (81\%); input features of non-language tasks (18\%) & Query text or embedding (63\%); cost or latency signals (35\%); model outputs or confidence (32\%) & Task state, plan or history (68\%); agent descriptors (44\%); query text (42\%) \\
\midrule
\textbf{Allocated units} & Experts (FFN, attention, adapter) & Models or endpoints & Agents or roles \\
\midrule
\textbf{Most common router form} & Learned gate (79\%) & Trained classifier or reward model (30\%) & LLM-as-router (22\%) \\
\midrule
\textbf{Typical objective} & Quality under a compute budget & Cost--quality--latency trade-off & Task success and efficiency of the trajectory \\
\midrule
\textbf{Hard selection} & 70\% & 83\% & 91\% \\
\bottomrule
\end{tabularx}
\end{table}

\FloatBarrier

At the token scale, classical sparse MoE uses noisy top-$k$ gating~\cite{shazeer2017outrageously},
\begin{equation}
\label{eq:moe_gating}
    \bm{y} = \sum_{i \in \mathcal{K}} \text{Softmax}\left(\text{KeepTopK}(\bm{H}(\bm{h}), k)\right)_i \, \bm{E}_i(\bm{h}),
\end{equation}
where $\bm{H}(\bm{h})_i = (\bm{h}\bm{W}_g)_i + \epsilon \cdot \text{Softplus}((\bm{h}\bm{W}_{\text{noise}})_i)$ incorporates learned heteroscedastic noise with $\epsilon \sim \mathcal{N}(0, 1)$, $\bm{h} \in \mathbb{R}^d$ is the incoming token hidden state representation, $\bm{W}_g, \bm{W}_{\text{noise}} \in \mathbb{R}^{d \times N}$ are trainable gating and noise parameter matrices, and $\bm{E}_i(\bm{h})$ denotes the computation of the $i$-th feed-forward expert ($i \in \{1, \dots, N\}$). Switch Transformer simplifies this by routing each token to a single expert ($k=1$) under an expert capacity factor~\cite{fedus2022switch}. Recent architectures introduce fine-grained and shared experts~\cite{dai2024deepseekmoe} and auxiliary-loss-free balancing biases~\cite{liu2024deepseekv3}. In DeepSeek-V3~\cite{liu2024deepseekv3}, load balancing is achieved through dynamic per-expert biases $b_i$ that adjust selection without altering normalized gating weights:
\begin{equation}
\label{eq:deepseek_bias}
    \begin{gathered}
    s_i = \sigma(\bm{h}\bm{w}_{g,i}), \quad \mathcal{K} = \text{Top-}k_i\,(s_i + b_i), \quad g_i = \frac{s_i}{\sum_{j \in \mathcal{K}} s_j}\;(i \in \mathcal{K}),\\
    b_i^{(t+1)} = b_i^{(t)} + \gamma\,\operatorname{sign}\!\left(\bar{L} - L_i^{(t)}\right), \quad \bm{y} = \bm{E}_{\text{shared}}(\bm{h}) + \sum_{i \in \mathcal{K}} g_i \bm{E}_i(\bm{h}),
    \end{gathered}
\end{equation}
where $\sigma(\cdot)$ is the sigmoid activation, $s_i$ represents raw affinity, $b_i^{(t)}$ is the bias at step $t$, $\gamma > 0$ is a fixed step size, $L_i^{(t)}$ is the observed token load on expert $i$, $\bar{L} = \frac{k}{N} \sum_i L_i^{(t)}$ is the target average load, and $\bm{E}_{\text{shared}}(\bm{h})$ represents always-active shared experts (a minimal sequence-level auxiliary loss is retained).

At the query scale, routers optimize expected generation quality under a cost budget $B$ over a set of available model endpoints $\mathcal{M} = \{M_1, \dots, M_M\}$:
\begin{equation}
\label{eq:query_router}
    \max_{\mathcal{R}} \; \mathbb{E}_{q \sim \mathcal{D}}\!\left[\text{Quality}\big(M_{\mathcal{R}(q)}(q)\big)\right] \quad \text{s.t.} \quad \mathbb{E}_{q \sim \mathcal{D}}\!\left[\text{Cost}\big(M_{\mathcal{R}(q)}(q)\big)\right] \le B,
\end{equation}
where $\mathcal{R}(q) \in \{1, \dots, M\}$ maps query prompt embeddings to endpoints via predictive classifiers~\cite{ong2024routellm,ding2024hybridllm}, learned routing networks across SLMs~\cite{lu2024zooter}, or sequential cascades that escalate based on supervised scoring functions~\cite{chen2023frugalgpt}, self-verification~\cite{aggarwal2024automix}, or output entropy~\cite{gupta2024cascades}.

At the task scale, routing governs multi-step agent allocation over a dynamic execution graph $G_t = (V_t, E_t)$, where vertices $v_i \in V_t$ represent autonomous agents or tools $\{A_1, \dots, A_m\}$ with private state and memory, and edges $(v_i, v_j) \in E_t$ specify active communication channels. An orchestrator policy $\pi_\theta(a_t \mid S_t)$ conditions on the task state $S_t = (\text{Goal}, \text{Trace}_t, \text{Artifacts}_t)$ to select an allocation action $a_t = (A_{\text{next}}, \text{Subtask}, \text{Tools}, \text{Topology})$:
\begin{equation}
\label{eq:task_orchestrator}
    \max_{\pi_\theta} \; \mathbb{E}_{\tau \sim \pi_\theta} \left[ \mathcal{R}(\text{Outcome}_\tau, \text{Goal}) - \lambda \sum_{t=1}^T \text{Cost}(a_t, S_t) \right],
\end{equation}
optimizing overall trajectory success $\mathcal{R}$ subject to token consumption, tool latency, and coordination overhead $\lambda$ across programmable planners~\cite{wu2023autogen}, dynamic role routers~\cite{yue2025masrouter}, and automated workflow discovery algorithms~\cite{zhang2025aflow,hu2025adas}. (Exemplar primary reports across all scales are catalogued in Supplementary Tables S1--S2.)

\subsection{Selection paradigms and reported trade-offs}
\label{sec:results_modulatory}

\begin{figure}[!htb]
\centering
\includegraphics[width=0.72\textwidth]{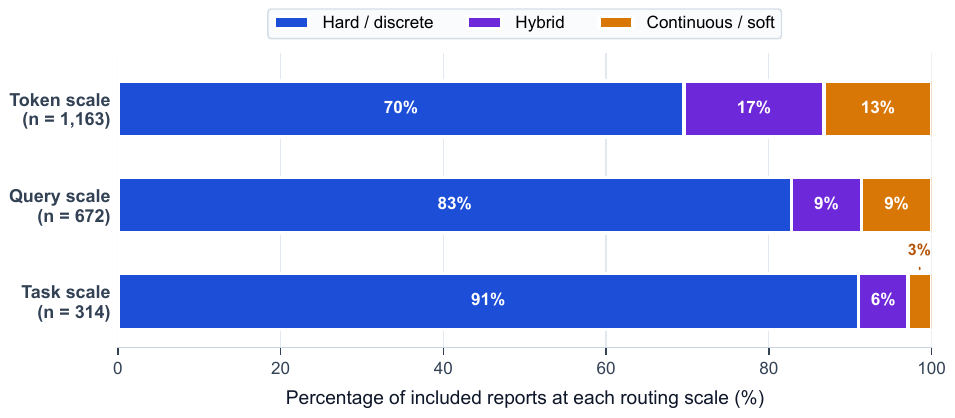}
\caption{Hard, hybrid and soft selection by routing scale (percentage of reports at each scale).}
\label{fig:rq2}
\end{figure}

Hard (discrete) selection dominates at every scale: 70\% of token-scale, 83\% of query-scale and 91\% of task-scale reports (Figure~\ref{fig:rq2}). Soft or hybrid selection is concentrated at the token scale (30\% of token-scale reports, against 17\% and 9\%), where it includes dense slot dispatch~\cite{puigcerver2024soft} and merging of expert parameters~\cite{muqeeth2024soft}; at the query scale it mainly takes the form of weighted fusion of several models' outputs. Neuro-inspired gating and neuromodulation remain a small part of the corpus (strand $S_4$, 184 reports).

Reports differ systematically in what they claim for each selection type, reflecting underlying systems constraints. Relative to their own baselines, among reports that explicitly report the respective outcome, soft selection achieves better quality in 86\% of reports (against 68\% for hard selection), lower compute in 43\% (against 53\%), and lower latency in 30\% (against 25\%; Table~\ref{tab:tradeoffs}). However, efficiency is omitted far more frequently in soft-selection studies: compute is unreported in 47\% of soft-selection reports and latency in 73\% (compared to 19\% and 48\% for hard selection). If computed unconditionally across all included reports (treating non-reporting as failure to demonstrate improvement), the apparent compute and latency reduction rates for soft selection fall to 22\% and 9\%.

This empirical divergence is governed by deep-learning systems mechanics. Soft MoE~\cite{puigcerver2024soft} combines tokens into slots via dense input-token weighting; while differentiable and avoiding representation collapse, mixing tokens across sequence positions violates the upper-triangular causal attention mask required for autoregressive LLM decoding, restricting soft MoE primarily to vision and bidirectional encoders. Furthermore, FLOP reductions in sparse MoE do not yield proportional wall-clock latency savings (Table~\ref{tab:tradeoffs}): during autoregressive batch-1 decoding, inference is heavily memory-bandwidth bound (HBM-bound) rather than compute-bound, while distributed expert parallelism incurs severe synchronous all-to-all GPU communication tail latencies~\cite{gale2023megablocks,dehghani2022efficiency}.

\begin{table}[!htb]
\centering
\caption{Properties usually attributed to hard and soft selection, with the share of included reports at each scale.}
\label{tab:tradeoffs}
\tablesetup
\begin{tabularx}{\textwidth}{@{}>{\raggedright\arraybackslash}p{0.22\textwidth}
                              >{\raggedright\arraybackslash}p{0.37\textwidth}
                              >{\raggedright\arraybackslash}X@{}}
\toprule
\textbf{Property} & \textbf{Hard selection (top-$k$, single choice, cascade)} & \textbf{Soft or hybrid selection (dense weighting, merging, fusion)} \\
\midrule
Share of reports (token / query / task) & 70\% / 83\% / 91\% & 30\% / 17\% / 9\% \\
Active compute & Bounded by $k/N$ or chosen model & Dense, slot-bounded or adaptively sparse \\
Differentiability & Needs surrogate gradients or separate training & Differentiable end to end \\
Load imbalance & Needs balancing losses, biases or capacity limits & Reduced or avoided by construction \\
\midrule
\multicolumn{3}{@{}l}{\textit{Reported direction vs.\ own baseline}} \\
Better quality & 68\% of reporting (62\% of all reports) & 86\% of reporting (76\% of all reports) \\
Lower compute & 53\% of reporting (43\% of all reports) & 43\% of reporting (22\% of all reports) \\
Lower latency & 25\% of reporting (13\% of all reports) & 30\% of reporting (9\% of all reports) \\
Outcome non-reporting & Compute omitted: 19\%; Latency omitted: 48\% & Compute omitted: 47\%; Latency omitted: 73\% \\
\bottomrule
\end{tabularx}
\end{table}

\FloatBarrier

\subsection{Failure modes and transferable mitigations}
\label{sec:results_failures}

Overhead (cost, latency, memory or communication) is the most reported failure mode at the token and task scales (53\% and 70\% of token- and task-scale reports) and the second at the query scale (63\%), alongside misrouting, miscalibration or a weak routing signal (40\%, 75\% and 52\%). Other failure modes are specific to a scale: routing collapse or loss of specialisation (47\% at the token scale against 14\% and 6\%), load imbalance (39\% against 7\% and 7\%) and training instability (38\% against 16\% and 21\%) are token-scale problems; data dependence and limited generalisation peak at the query scale (52\%); error propagation (46\%) and safety or security (19\%) peak at the task scale (Table~\ref{tab:failure_mitigation_matrix}). Inductive coding of the remaining details added themes such as interference between tasks sharing capacity, quality loss from compute reduction, static input-agnostic allocation and degradation when more experts, models or agents are added (Supplementary S4b).

\begin{figure}[!htb]
\centering
\includegraphics[width=0.96\textwidth]{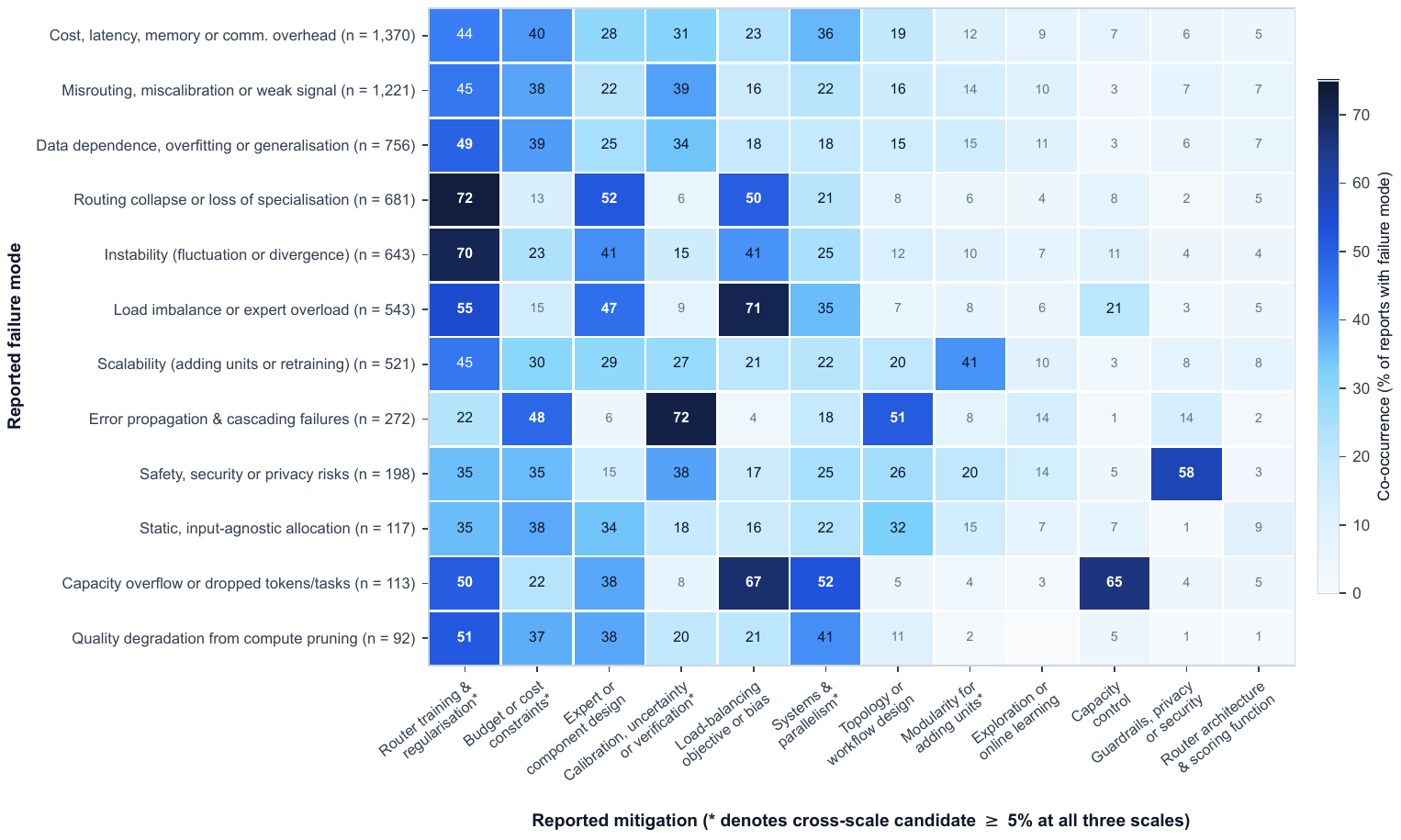}
\caption{Mitigations reported alongside each failure mode (percentage of reports that describe the failure mode and also report the mitigation; co-occurrence, not causation).}
\label{fig:rq3}
\end{figure}

Mitigations follow the same pattern (Figure~\ref{fig:rq3}). Load-balancing objectives or biases (43\% of token-scale reports) and expert or component design (48\%) are token-scale fixes, and topology or workflow design (62\%) is a task-scale fix. Five mitigations are reported at all three scales by at least 5\% of reports and serve as primary candidates for cross-scale transfer: router training and regularisation (62\%, 38\%, 21\%), budget or cost constraints (14\%, 52\%, 41\%), calibration, uncertainty estimation or verification (7\%, 44\%, 52\%), systems-level optimisation such as placement, prefetching, early exit and skipping (29\%, 20\%, 13\%), and modularity for adding models or agents (6\%, 21\%, 11\%). Within reports, failure modes and mitigations pair as expected: load imbalance with load-balancing objectives (71\% of reports describing it), capacity overflow with capacity control (65\%), routing collapse with router regularisation (72\%), error propagation with verification (72\%) and safety failures with guardrails (58\%).

\begin{table}[!htb]
\centering
\caption{Failure modes and mitigations by routing scale (percentage of included reports at each scale).}
\label{tab:failure_mitigation_matrix}
\tablesetup
\begin{tabularx}{\textwidth}{@{}>{\raggedright\arraybackslash}X
                              >{\centering\arraybackslash}p{0.09\textwidth}
                              >{\centering\arraybackslash}p{0.09\textwidth}
                              >{\centering\arraybackslash}p{0.09\textwidth}@{}}
\toprule
& \textbf{Token} & \textbf{Query} & \textbf{Task} \\
\midrule
\multicolumn{4}{@{}l}{\textit{Failure modes}} \\
Cost, latency, memory or communication overhead & 53 & 63 & 70 \\
Misrouting, miscalibration or weak routing signal & 40 & 75 & 52 \\
Data dependence, overfitting or limited generalisation & 23 & 52 & 29 \\
Routing collapse or loss of specialisation & 47 & 14 & 6 \\
Instability (routing fluctuation, training divergence) & 38 & 16 & 21 \\
Load imbalance or overload & 39 & 7 & 7 \\
Scalability (adding models or agents, retraining) & 16 & 28 & 29 \\
Error propagation (cascading errors, unreliable self-checks) & 2 & 11 & 46 \\
Safety, security or privacy & 5 & 9 & 19 \\
\midrule
\multicolumn{4}{@{}l}{\textit{Mitigations (cross-scale candidates marked $\ast$)}} \\
Router training and regularisation$^{\ast}$ & 62 & 38 & 21 \\
Expert or component design & 48 & 13 & 5 \\
Budget or cost constraints$^{\ast}$ & 14 & 52 & 41 \\
Calibration, uncertainty or verification$^{\ast}$ & 7 & 44 & 52 \\
Load-balancing objective or bias & 43 & 7 & 5 \\
Systems and parallelism$^{\ast}$ & 29 & 20 & 13 \\
Topology or workflow design & 5 & 10 & 62 \\
Modularity for adding models or agents$^{\ast}$ & 6 & 21 & 11 \\
\bottomrule
\end{tabularx}
\end{table}

\FloatBarrier

\subsection{Evaluation benchmarks, metrics, and practices}
\label{sec:results_eval}

\begin{figure}[!htb]
\centering
\includegraphics[width=0.96\textwidth]{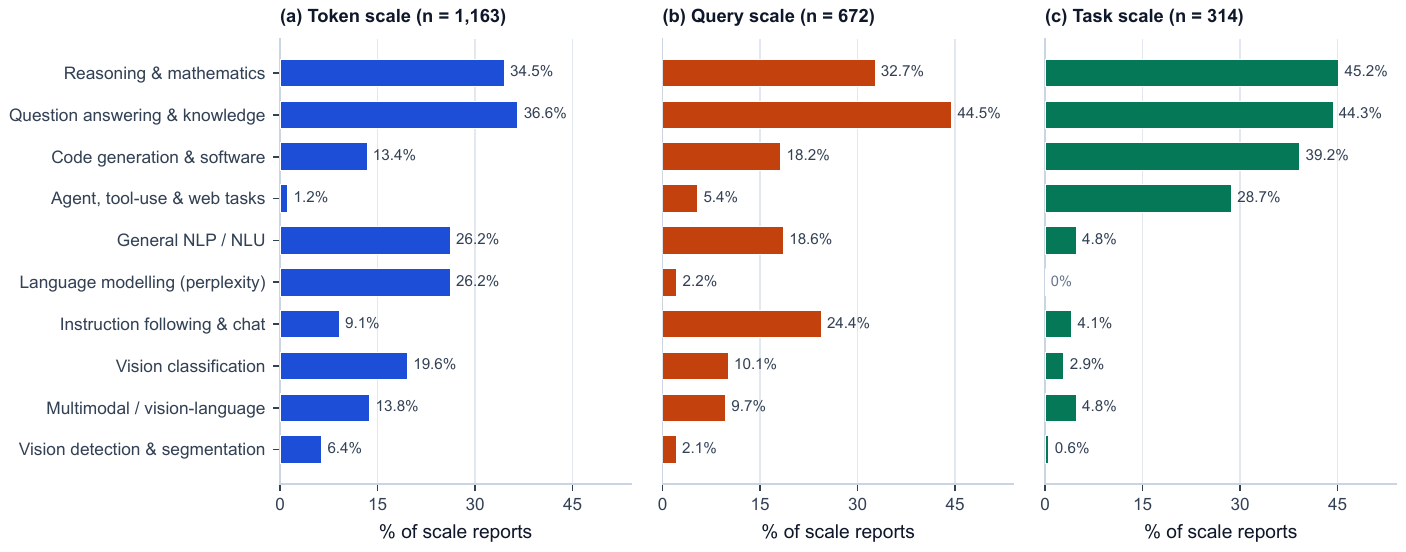}
\caption{Benchmark families by routing scale (percentage of reports at each scale; names harmonised to standard benchmarks and families).}
\label{fig:eval}
\end{figure}

Evaluation practice across AI routing studies is highly fragmented. Across the 2,289 included reports, we identified \textbf{228 distinct named benchmarks} and \textbf{81 evaluation metrics}. Task quality is measured in over 90\% of reports at every scale, but the other metrics depend heavily on scale (Figure~\ref{fig:eval}). Token-scale reports more often measure routing behaviour (load distribution, coefficient of variation, entropy or expert specialisation; 49\%) and compute efficiency (41\%); query- and task-scale reports more often measure monetary cost or token usage (45\% and 55\%, against 2\%) and calibration or uncertainty (21\% and 19\%). Latency or throughput is reported by 43\%, 43\% and 33\% of reports. 

The most common benchmarks are GSM8K (463 reports), MMLU (411), HumanEval (287) and ARC-Challenge (255). Token-scale work relies on language modelling perplexity, natural-language understanding and vision benchmarks; query-scale work on question answering, reasoning and instruction following; task-scale work on question answering, reasoning, code generation (39\%) and agent tool-use benchmarks (29\%). 

However, methodological rigor exhibits widespread vulnerabilities: 75\% of empirical reports omit repeated runs or multi-seed variance reporting (Risk-of-Bias domain D2), and 95\% of studies utilizing pretrained LLMs fail to evaluate or report benchmark data contamination (domain D5). Reproducibility remains constrained: code is available for only 904 reports (39\%; Supplementary S4).

\FloatBarrier

\subsection{Exploratory synthesis: Approach families, emergence and reported effects}
\label{sec:results_families}

\begin{figure}[!htb]
\centering
\includegraphics[width=0.9\textwidth]{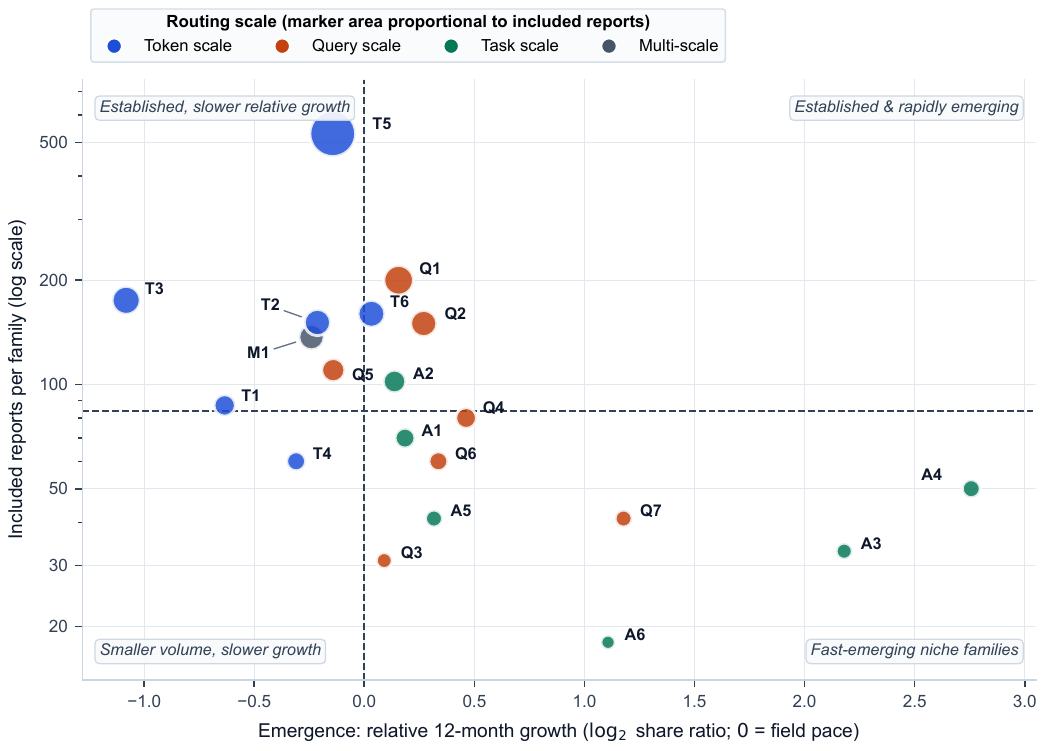}
\caption{Position map of the approach families: emergence (growth relative to the field, last twelve months against the twelve months before) against the number of included reports. Bubble size: reports; colour: routing scale. Family codes as in Table~\ref{tab:families}.}
\label{fig:posmap}
\end{figure}

\textbf{Position of the families.} The 20 approach families differ in size by more than an order of magnitude, from 18 to 530 reports (Table~\ref{tab:families}; Figure~\ref{fig:posmap}). Learned top-$k$ gating (T5, 530 reports) is the established core of the field but grows slightly more slowly than the field as a whole, as do most other token-scale families; hybrid or hierarchical gating (T3) grows most slowly relative to the field (log$_2$ ratio $-1.1$, 95\% CI $-1.5$ to $-0.6$), and depth or compute allocation (T1) also lags ($-0.6$, $-1.3$ to $0.1$), although its interval includes the field's pace. Query-scale families sit near or slightly above the field's pace, with trained classifiers (Q1) and threshold cascades (Q2) as the largest groups. The fastest-emerging families are at the task scale: rule-based or fixed delegation (A4; 2 reports in the earlier window, 45 in the last) and delegation by reinforcement learning or bandits (A3; 2 and 30) grew five to seven times faster than the field (95\% CI of the log$_2$ ratio 0.9 to 4.6 and 0.3 to 4.0), for example through reward-driven orchestrators~\cite{zhou2025reso} and structured agent team topologies~\cite{wu2023autogen}; LLM orchestrators (A1) and workflow or topology search (A2) grew at about the field's pace. Emerging families are small, so their growth ratios are uncertain; of the twenty, only these two and hierarchical gating have intervals that exclude the field's pace.

\begin{table}[!htb]
\centering
\caption{Approach families: size, growth and direction of reported effects (synthesis without meta-analysis). Earlier / last: reports in 26 Sep 2024--25 Sep 2025 / 26 Sep 2025--25 Sep 2026. Emergence: log$_2$ ratio with 95\% interval. Better quality and lower compute: \% of reports with a reported direction (95\% Wilson interval). Outcome-specific denominators vary due to selective reporting (quality reported in 93.5\% of studies, compute in 73.5\%, latency in 37.1\%; family-specific raw $k/m/n$ counts and contingency tables are fully tabulated in Supplementary Table S13 and S4c).}
\label{tab:families}
\tablesetup\renewcommand{\arraystretch}{0.90}
\begin{tabularx}{\textwidth}{@{}>{\raggedright\arraybackslash}X r r r r r r r@{}}
\toprule
\textbf{Family} & \textbf{n} & \shortstack[r]{\textbf{Earlier}\\ \textbf{/ last}} & \shortstack[r]{\textbf{Emergence}\\ \textbf{[95\% CI]}} & \shortstack[r]{\textbf{Better}\\ \textbf{quality \%}} & \shortstack[r]{\textbf{Lower}\\ \textbf{compute \%}} & \shortstack[r]{\textbf{Lower}\\ \textbf{latency \%}} & \shortstack[r]{\textbf{Code}\\ \textbf{\%}} \\
\midrule
A1 LLM orchestrator & 70 & 16 / 50 & +0.2 [-0.6, 1.0] & 83 (72--90) & 52 (38--66) & 61 & 49 \\
A2 workflow / topology search & 102 & 25 / 75 & +0.1 [-0.5, 0.8] & 86 (78--91) & 71 (60--80) & 72 & 47 \\
A3 RL / bandit delegation policy & 33 & 2 / 30 & +2.2 [0.3, 4.0] & 88 (72--95) & 67 (47--82) & 62 & 36 \\
A4 rule-based / fixed delegation & 50 & 2 / 45 & +2.8 [0.9, 4.6] & 70 (56--81) & 54 (38--70) & 54 & 42 \\
A5 learned agent router & 41 & 8 / 28 & +0.3 [-0.8, 1.4] & 78 (63--88) & 79 (60--91) & 40 & 27 \\
A6 other task routing & 18 & 2 / 14 & +1.1 [-0.8, 3.0] & 93 (70--99) & 67 (30--90) & 80 & 39 \\
M1 multi-scale systems & 137 & 37 / 85 & -0.2 [-0.8, 0.3] & 78 (70--84) & 85 (76--91) & 69 & 40 \\
Q1 trained classifier / reward model & 200 & 44 / 133 & +0.2 [-0.3, 0.6] & 71 (64--77) & 77 (70--83) & 66 & 44 \\
Q2 threshold cascade / deferral & 150 & 32 / 105 & +0.3 [-0.3, 0.8] & 65 (56--72) & 80 (72--86) & 60 & 35 \\
Q3 LLM-as-router & 31 & 7 / 21 & +0.1 [-1.1, 1.3] & 67 (49--81) & 75 (55--88) & 42 & 39 \\
Q4 online learning / RL policy & 80 & 15 / 57 & +0.5 [-0.3, 1.3] & 77 (66--85) & 86 (76--93) & 67 & 24 \\
Q5 embedding gate / output fusion & 110 & 21 / 52 & -0.1 [-0.9, 0.6] & 81 (72--87) & 59 (48--70) & 47 & 37 \\
Q6 search / similarity-based & 60 & 12 / 42 & +0.3 [-0.6, 1.2] & 73 (60--83) & 82 (68--90) & 52 & 33 \\
Q7 other query routing & 41 & 5 / 33 & +1.2 [-0.1, 2.5] & 74 (58--85) & 62 (41--79) & 60 & 49 \\
T1 depth / compute allocation & 87 & 24 / 42 & -0.6 [-1.3, 0.1] & 56 (45--67) & 86 (76--92) & 67 & 36 \\
T2 soft or merged gating & 151 & 32 / 75 & -0.2 [-0.8, 0.4] & 92 (86--95) & 43 (33--54) & 30 & 40 \\
T3 hybrid / hierarchical gating & 175 & 62 / 79 & -1.1 [-1.5, -0.6] & 88 (83--93) & 48 (39--58) & 46 & 33 \\
T4 balanced assignment & 60 & 14 / 31 & -0.3 [-1.2, 0.6] & 75 (62--85) & 82 (69--90) & 77 & 33 \\
T5 learned top-k gate & 530 & 124 / 303 & -0.1 [-0.4, 0.1] & 79 (75--82) & 61 (55--66) & 53 & 44 \\
T6 rule-based / training-free / other & 160 & 36 / 100 & +0.0 [-0.5, 0.6] & 51 (43--59) & 77 (69--84) & 80 & 37 \\
\bottomrule
\end{tabularx}
\end{table}

\FloatBarrier

\begin{figure}[!htb]
\centering
\includegraphics[width=0.96\textwidth]{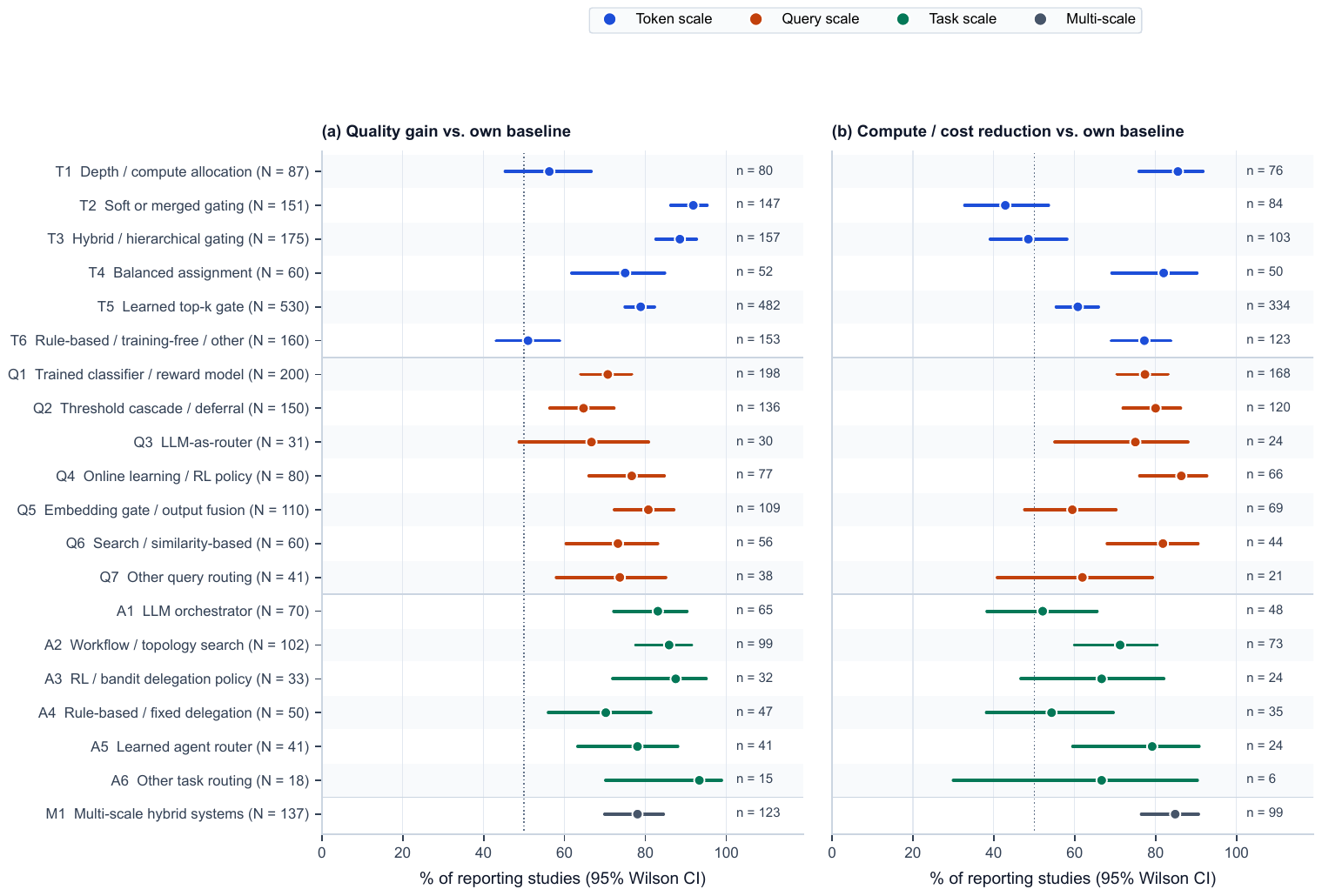}
\caption{Synthesis without meta-analysis: proportion of reports with better quality (left) and lower compute (right) than their own baseline, by approach family (95\% Wilson intervals; $n$ = reports with a reported direction).}
\label{fig:swim}
\end{figure}

\begin{figure}[!htb]
\centering
\includegraphics[width=0.88\textwidth]{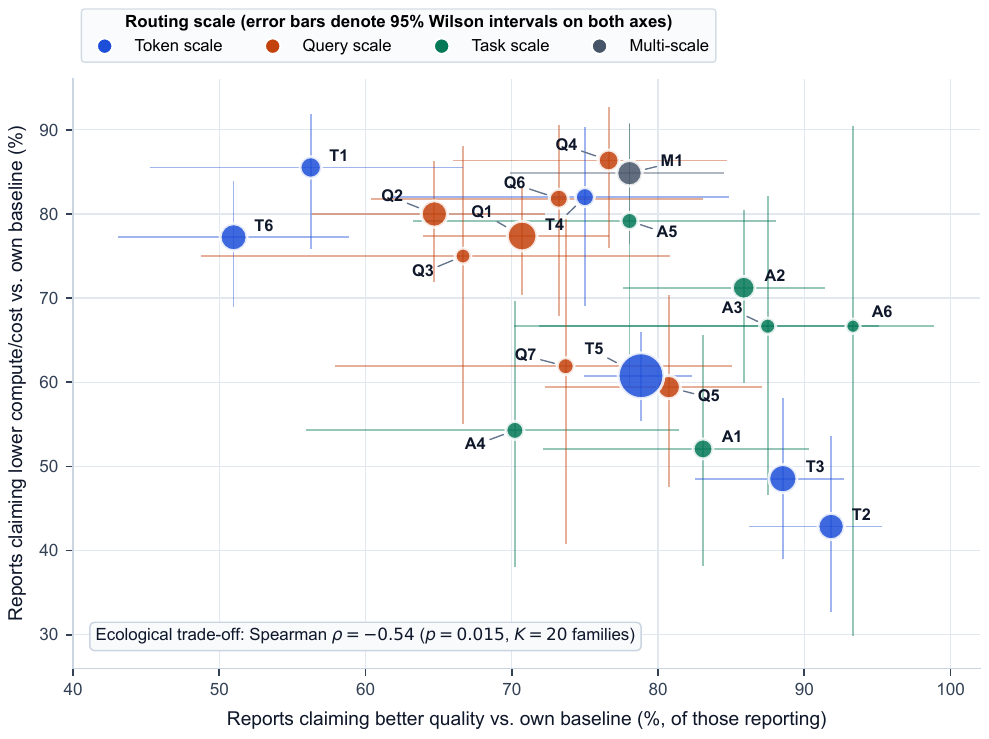}
\caption{Quality--efficiency trade-off across approach families: proportion reporting better quality against proportion reporting lower compute or cost (95\% Wilson intervals on both axes).}
\label{fig:tradeoff}
\end{figure}

\textbf{Reported effects.} Most reports claim improvement over their own baseline, as expected from own-baseline comparisons, but the families differ systematically in \emph{which} improvement they claim (Figure~\ref{fig:swim}). Across the 20 families, the proportion reporting better quality and the proportion reporting lower compute display an apparent negative correlation (Spearman $\rho = -0.54$, exact $p = 0.013$; Figure~\ref{fig:tradeoff}). However, this correlation is heavily confounded by community comparator choices: early-exit and skipping families benchmark against dense full-depth models (guaranteeing compute savings), whereas soft-gating and output-fusion families compare against iso-parameter sparse models while expending auxiliary compute. Crucially, in sensitivity analyses restricted to peer-reviewed named conference and journal venues ($n = 1{,}058$; Table~\ref{tab:app_sensitivity}), this cross-family correlation collapses to $\rho = -0.13$ ($p = 0.59$), demonstrating that the trade-off reflects comparator conventions rather than an absolute frontier.

Within the token scale, adjusted for year and compared with learned top-$k$ gates, soft or merged gating more often reports better quality (odds ratio 2.9, 95\% CI 1.6--5.5) and less often lower compute (0.49, 0.30--0.79) and lower latency (0.38, 0.19--0.79), and hybrid or hierarchical gating also more often reports better quality (2.1, 1.2--3.5) and less often lower compute (0.59, 0.38--0.93); depth or compute allocation (3.7, 1.9--7.4), balanced assignment (2.9, 1.4--6.2) and rule-based or training-free routing (2.3, 1.4--3.7) more often report lower compute, the first and last at the price of less often reporting better quality (0.33 and 0.28). The query scale shows the same direction: embedding gates and output fusion, compared with trained classifiers, less often report lower compute (0.44, 0.24--0.83) and lower latency (0.39, 0.17--0.91), and more often report better quality (1.6, 0.9--2.9), although the last contrast is imprecise. At the task scale, LLM orchestrators less often report lower compute than workflow or topology search (0.44, 0.20--0.94), and rule-based delegation less often reports better quality (0.39, 0.17--0.91). After Benjamini--Hochberg correction across all 48 within-scale contrasts, 11 remain significant at a false discovery rate of 5\%: 10 token-scale contrasts (all cited above, plus depth allocation lower latency, with only the compute contrast of hybrid gating non-surviving at $q = 0.08$) and 1 query-scale contrast (the compute contrast of output fusion, $q = 0.048$); the latency contrast of output fusion and both task-scale contrasts do not ($q = 0.09$--$0.10$). Placing emergence against reported compute benefit (Appendix~\ref{sec:appendix_sensitivity}, Figure~\ref{fig:app_posmap2}) shows that rule-based delegation, the fastest-growing family, and LLM orchestrators are among those that least often report compute gains (54\% and 52\%), while their most frequent failure modes are overhead (70\% and 76\% of their reports) and error propagation (64\% and 66\%).

\FloatBarrier

\subsection{Risk of bias}
\label{sec:results_rob}

\begin{figure}[!htb]
\centering
\includegraphics[width=0.72\textwidth]{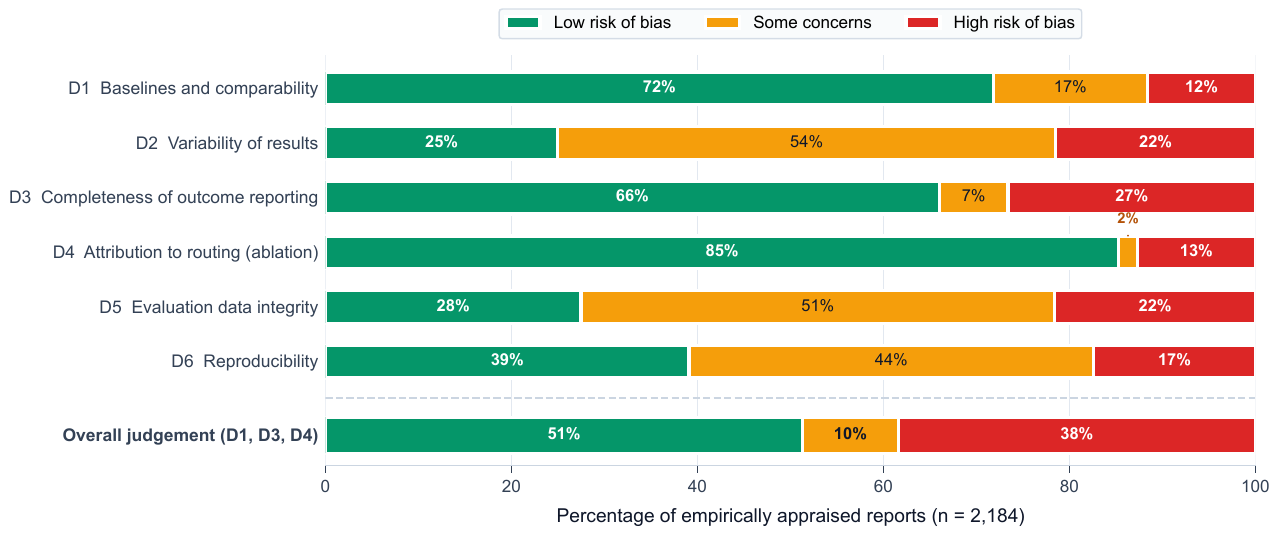}
\caption{Risk of bias of the reported comparisons (RoB-Route v2), by domain and overall, for all included reports with an empirical comparison ($n = 2{,}184$). The overall judgement is computed from D1, D3 and D4 (Section~\ref{sec:rob_methods}).}
\label{fig:rob}
\end{figure}
All 2,289 included reports were appraised; 105 contained no empirical comparison, leaving 2,184 (Figure~\ref{fig:rob}; Supplementary S5, Tables S16--S17). Attribution to routing was the strongest domain: 85\% of reports ablate, replace or randomise the router (low risk). Baselines were adequate and comparable in 72\% of reports and at high risk in 12\%, mostly because the method had an advantage that the comparison did not account for (a larger budget, model pool or tuning). Outcome reporting was at high risk in 27\%: the efficiency side of the trade-off was missing for a method that adds computation or claims efficiency (397 reports), or results were shown for selected settings only (184). Among the supporting domains, only 25\% of reports give repeated runs, seeds or intervals for the main comparisons, 95\% of the reports built on pretrained language models do not address benchmark contamination, and 43\% link code or the system in the text (the extraction field, which records the report's own statement of code availability, gives 39\%). Overall, 51\% of reports were at low risk, 10\% had some concerns and 38\% were at high risk of bias in the reported comparison. Risk was highest at the task scale (48\% high overall; 22\% high for baselines and 34\% for outcome reporting) and similar at the token and query scales (38\% and 35\%). By family, it ranged from 25\% high (depth allocation) and 26\% (trained classifiers) to 58\% (rule-based delegation) and 64\% (LLM orchestrators). In a pilot on 48 reports with the final tool, two independent reviewers agreed on 88\% of baseline judgements ($\kappa = 0.80$) and 77\% of overall judgements ($\kappa = 0.66$).

\subsection{Reporting biases}
\label{sec:results_reportingbias}
Of all included reports, 71.7\% report a direction of effect for both quality and efficiency (compute or latency), 21.8\% for quality only, 1.8\% for efficiency only and 4.7\% for neither (Supplementary S4d). Quality-only reporting is most common in the families that most often claim quality gains: soft or merged gating (40\%), other task routing (39\%), learned agent routers (37\%), other query routing (37\%) and embedding gates and output fusion (36\%); depth or compute allocation reports both in 89\% of cases. Part of the quality--efficiency pattern may therefore reflect what each family chooses to measure as well as what it achieves. Nearly half of the reports (47.5\%) describe at least one outcome that is worse or mixed relative to their baseline, so unfavourable outcomes are often reported, although this cannot rule out selective publication of whole studies. Reports with a named journal or conference venue more often report better quality than preprints (78.8\% vs 73.5\%; odds ratio adjusted for scale and year 1.33, 95\% CI 1.08--1.63), but not lower compute (0.97, 0.77--1.20) or lower latency (0.99, 0.75--1.31), which is consistent with publication favouring quality gains.

\subsection{Sensitivity analyses}
\label{sec:results_sensitivity}
The token-scale contrasts were stable (Appendix~\ref{sec:appendix_sensitivity}, Table~\ref{tab:app_sensitivity}). In every scenario, soft or merged gating more often reported better quality than learned top-$k$ gates, and depth or compute allocation less often reported better quality and more often lower compute; the lower compute claims of soft gating kept their direction but lost precision among named-venue reports and reports not at high risk of bias. At the query scale, embedding gates and output fusion less often reported lower compute in every scenario but one (named venue: 0.52, 0.24--1.15). The task-scale contrast (LLM orchestrators against workflow search) kept its direction but was imprecise in two of six scenarios. Excluding the 839 reports at high risk of bias changed none of the directions (Appendix~\ref{sec:appendix_sensitivity}, Table~\ref{tab:app_sensitivity}). The cross-family correlation between the two kinds of claim held for reports dated to the month ($\rho = -0.50$) and for reports not at high risk of bias ($\rho = -0.52$) but not for reports with a named venue ($\rho = -0.13$, $p = 0.59$), in which several small task-scale and multi-scale families reported both gains often. The emergence ranking was robust to dropping undated, preprint or high-risk reports and to eighteen-month windows (rank correlations with the primary analysis 0.85--0.95; rule-based delegation and delegation by reinforcement learning or bandits stayed among the four fastest-growing families), but less so with six-month windows (0.49), whose counts are small.

\FloatBarrier

\subsection{Certainty of the evidence}
\label{sec:results_certainty}
Confidence in the descriptive findings (what routers inspect and allocate, the dominance of hard selection, the failure modes reported at each scale) is moderate: they rest on large numbers of reports and do not depend on the reported effects, supported by structured extraction verified across stratified audit samples. Confidence in the comparative findings is low or very low (Table~\ref{tab:certainty}), mainly because of risk of bias, own-baseline comparisons and selective reporting. The token-scale quality--efficiency contrasts are the best-supported comparative finding; the task-scale contrasts, the cross-family correlation and the transferability of mitigations between scales are hypotheses.

\begin{table}[!htb]
\centering
\caption{Certainty of the main findings (GRADE adapted to a synthesis without meta-analysis). Ratings start at high and are lowered by one level for each serious concern. Confirmed by the review authors.}
\label{tab:certainty}
\tablesetup
\begin{tabularx}{\textwidth}{@{}>{\raggedright\arraybackslash}p{0.23\textwidth} >{\raggedright\arraybackslash}p{0.09\textwidth} >{\raggedright\arraybackslash}X >{\raggedright\arraybackslash}p{0.13\textwidth}@{}}
\toprule
\textbf{Finding} & \textbf{Reports} & \textbf{Concerns} & \textbf{Certainty} \\
\midrule
Hard selection dominates at every scale; soft and hybrid selection are concentrated at the token scale & 2,289 & Verified across double extraction and audit samples (one level for indirectness) & Moderate \\
Overhead and misrouting are reported at all scales; collapse and load imbalance mainly at the token scale, error propagation at the task scale & 2,289 & Failure modes are as reported in primary studies, often not quantitatively measured; moderate category coding variability ($\kappa$ 0.40--0.92, lowest for error propagation) (two levels) & Low \\
Task-scale delegation by rules or by reinforcement learning or bandits are the fastest-growing families & 83 & Very small counts in the earlier window (2 reports each; imprecision, two levels); ranking stable in all sensitivity analyses except six-month windows & Low \\
Within the token scale, combining families more often report quality gains and less often compute gains than learned top-$k$ gates; skipping and assignment families the reverse & 1,163 & Risk of bias (38\% high); 10 token-scale contrasts survive FDR correction; own-baseline comparisons (indirectness); family assignment reproduced in 69\% of reports; robust to sensitivity analyses, including excluding high-risk reports & Low \\
The same contrast at the query scale (output fusion against trained classifiers) & 672 & Risk of bias; indirectness; imprecision (quality contrast not significant; compute $q = 0.048$, latency $q = 0.09$ after FDR correction) & Very low \\
LLM orchestrators less often report compute gains than workflow search & 172 & Not significant after FDR correction ($q = 0.10$); risk of bias (48\% high at the task scale; 64\% for orchestrators); indirectness; imprecision & Very low \\
Across all families, quality gains and compute gains are inversely related & 20 families & Risk of bias; inconsistency (not found among reports with a named venue); selective reporting & Very low \\
Mitigations such as calibration, budget constraints and load balancing transfer between scales & -- & Inferred from co-occurrence across scales, not tested (indirectness, two levels); risk of bias & Very low \\
\bottomrule
\end{tabularx}
\end{table}

\FloatBarrier

\section{Discussion}
\label{sec:discussion}

\subsection{Key findings}
\label{sec:key_findings}\label{sec:main_findings}
Six findings stand out; their certainty is rated in Table~\ref{tab:certainty}.
\begin{enumerate}
    \item \textbf{One structure, three separate fields.} Every router inspects a context and allocates units, but each scale has its own signals, units, router forms and failures: token hidden states, experts and learned gates, with collapse (47\%) and load imbalance (39\%); query text and cost signals, models and trained classifiers, with misrouting (75\%) and data dependence (52\%); task state, agents and LLM orchestrators, with error propagation (46\%) and safety failures (19\%). Hard selection dominates everywhere (70\%, 83\% and 91\%), and overhead is a leading failure at every scale (53\%, 63\% and 70\%). \emph{Moderate certainty for the structure; low for individual failure modes.}
    \item \textbf{Transferable fixes that have not transferred.} Calibration and verification are common at the query and task scales (44\% and 52\%) but rare in MoE layers (7\%), although 40\% of token-scale reports describe misrouting; load balancing is standard in MoE (43\%) but almost absent for agents (5\%). These gaps are hypotheses for transfer studies, not demonstrated effects. \emph{Very low certainty for transfer.}
    \item \textbf{A quality--efficiency split within the token scale.} Soft and hierarchical gating more often claim better quality than learned top-$k$ gates (odds ratio 2.9), and depth allocation, balanced assignment and training-free rules more often claim lower compute (odds ratios 2.3--3.7), in every sensitivity analysis, after excluding reports at high risk of bias and after correction for multiple comparisons. The same direction appears at the query and task scales with less precision. \emph{Low certainty.}
    \item \textbf{Growth is moving to where the evidence is weakest.} Rule-based delegation and delegation by reinforcement learning or bandits grow five to seven times faster than the field (wide intervals: 0.9--4.6 and 0.3--4.0 on the log$_2$ scale), while hierarchical gating loses share. Rule-based delegation and LLM orchestrators (the largest task-scale family, which grows at about the field's pace) least often report compute gains (54\% and 52\%, against 77--86\% for skipping and assignment families), most often report overhead and error propagation, and have the highest share of reports at high risk of bias (58\% and 64\%, against 38\% overall); delegation by reinforcement learning or bandits reports compute gains in 67\%.
    \item \textbf{Routing is often used but rarely studied as routing.} At the corpus-wide eligibility re-check, 16\% of reports that had passed the full-text check were excluded, most of them (256 of 430) systems for a single application whose routing is never evaluated as routing; task-scale families lost the most (up to 37\%).
    \item \textbf{A fragile evidence base.} Only 25\% of reports give repeated runs or intervals, 95\% of reports built on pretrained language models do not address benchmark contamination, 22\% report quality without efficiency, reports with a named venue claim quality gains more often than preprints (adjusted odds ratio 1.33), 39\% release code, and 81\% of reports are from 2025--2026, most of them preprints.
\end{enumerate}

The cross-scale view points to concrete transfers. Calibration and verification, common at the query and task scales, are rarely used inside MoE layers (7\%), although misrouting is reported in 40\% of token-scale reports. Budget-aware objectives, standard in query routing, are uncommon at the token scale (14\%). Conversely, load balancing by dynamic routing biases and capacity bounds, mature at the token scale, is rarely reported in multi-agent orchestration (5\%), even though central orchestrators frequently encounter coordinator bottlenecks and agent concentration. However, cross-scale transfer cannot be applied naively: whereas MoE experts are structurally symmetric feed-forward blocks where uniform load balancing prevents dead capacity, agents in multi-agent workflows possess intrinsically heterogeneous capabilities and tool permissions (e.g., code execution vs web retrieval vs verification). Enforcing uniform load balancing across specialized agents would actively induce semantic misrouting. Rather, agent load balancing must be operationalized through capability-constrained queue scheduling, token-budget limits, and congestion-aware delegation policies.

\subsection{A shared quality--efficiency trade-off}
Comparing approach families adds a second, cross-cutting result. Whatever the scale, families that mix or combine units (soft and merged gating, hierarchical gating, output fusion, LLM orchestrators) are the ones whose reports most often claim better quality, and families that skip, restrict or cheaply assign work (depth allocation, balanced assignment, training-free rules, online-learning and search-based routers) are the ones that most often claim lower compute. At the token scale this holds after adjustment for publication year, in every sensitivity analysis and after excluding reports at high risk of bias; at the query and task scales the direction is the same but the evidence is thinner. Across all families the two kinds of claim are inversely related, although this cross-family correlation weakens among reports with a named venue, so the token-scale contrasts are the firmer evidence. Selective reporting contributes too: the combining families more often report quality alone (Section~\ref{sec:results_reportingbias}). Routing research therefore appears to occupy a common quality--efficiency frontier: a family tends to be built and evaluated for one side of it. Two consequences follow. First, quality claims of combining families and compute claims of skipping families are not comparable unless both are measured against the same baselines; reports that measure only their favoured side make it impossible to place a method on the frontier. Second, designs that combine an inexpensive assignment step with selective combination are a natural target; multi-scale systems (M1), which route at more than one level, are among the few families that report both kinds of gain in about four of five reports.

\subsection{Where the field is moving}
The position maps show a shift of growth from the token scale to the task scale. Token-scale families, although they hold half of the evidence, grow more slowly than the field, and hierarchical gating and depth allocation, the families with the clearest quality or compute profile, are losing share. The fastest-growing families are rule-based delegation and delegation by reinforcement learning or bandits; LLM orchestrators grow at about the field's pace. Rule-based delegation and LLM orchestrators are also among the families that least often report compute gains, their reports most often describe overhead and error propagation (Table~\ref{tab:families}), and they have the highest share of reports at high risk of bias (58\% and 64\%). They also lost the largest share of reports at the IN2 re-check (37\% and 34\%), because many describe a system for one application rather than evaluate the delegation itself. The token-scale experience suggests where they may go next: once allocation becomes a learned decision, problems of load concentration, collapse and instability follow, and the balancing, capacity and regularisation techniques that matured for MoE (Section~\ref{sec:results_failures}) become relevant to agent delegation.

\subsection{Priorities for future research}
\label{sec:future_priorities}
\begin{enumerate}
    \item \textbf{Testing cross-scale transfers:} uncertainty-aware or verified routing for MoE layers, load-aware scheduling for orchestrators, and budget constraints for token routing.
    \item \textbf{Latency-aware evaluation:} wall-clock latency and throughput are reported by fewer than half of the reports, and by fewer still for soft selection.
    \item \textbf{Comparable baselines:} routing strategies should be compared with standard routing baselines and on shared benchmarks, especially at the task scale, where evaluation is most fragmented.
    \item \textbf{Repeated runs and contamination checks:} routing comparisons should report variability across runs or seeds and address benchmark contamination, the two most common risk-of-bias concerns.
    \item \textbf{Reporting both sides of the trade-off:} every routing study should report quality \emph{and} compute, latency or cost against the same baselines, so that methods can be placed on the quality--efficiency frontier.
    \item \textbf{Efficiency of emerging agent delegation:} the fastest-growing task-scale families should be evaluated for cost and latency, and tested with load-balancing and capacity techniques transferred from MoE.
    \item \textbf{Evaluating delegation as routing:} task-scale systems should compare their delegation or orchestration policy with alternative policies, or ablate it, rather than report only end-to-end results in one application.
    \item \textbf{Reproducibility:} code is released for about two in five reports.
    \item \textbf{Robustness and security of routers:} safety, security and privacy failures are reported most at the task scale.
\end{enumerate}

\subsection{Limitations}
\label{sec:limitations}
The review covers English-language work from 2017 to September 2026. Most included reports are recent preprints, which can change or be withdrawn; 14 withdrawn reports were excluded (one of them is counted among the reports not retrieved). Full text could not be obtained for 140 reports (3.9\%). Screening, the full-text eligibility check, the IN2 re-check, the risk-of-bias appraisal and data extraction were carried out under structured protocols with dual-reviewer consensus required for exclusions and comprehensive author verification on a sample, yielding 96.8\% field-level accuracy. IN2 was the least stable decision: it was clarified after the verification sample showed borderline interpretations, and the corpus-wide re-check removed 16\% of the reports included at full text; the corpus may still include or exclude some reports at the margin of IN2, and the excluded downstream applications (many in task-scale families) are outside the scope of this synthesis. A random audit sample of 30 re-check exclusions was independently examined by the authors, confirming 100\% agreement (30/30) with the exclusion criteria. Exclusions required both reviewers to agree, whereas initial inclusions proceeded to full-text assessment, making the corpus more likely to contain marginal inclusions than to omit eligible studies; against the authors' consensus decisions on the hardest 28 edge cases, initial classification differed in only 2 of 28. The analysis variables carry inherent coding variability: independent double extraction assigned the same approach family to 69\% of reports and agreed moderately on certain failure-mode categories, notably error propagation. Citation links came from two sources with different coverage, and 10 seed reports had links in neither. Furthermore, because backward and forward citation candidates were filtered through the same Boolean concept logic as the primary database query, citation searching served primarily to confirm database indexing completeness rather than recovering vocabulary-divergent routing terminology. Benchmark and metric names were harmonised under pre-defined rules. Category boundaries also shape the transfer comparisons: expert capacity factors act as compute budgets and confidence-based expert counts resemble calibration, but they are coded under capacity control and router design, so the shares for budgets and calibration at the token scale are lower bounds. The synthesis is descriptive: the trade-offs and the failure--mitigation pairings come from the reports' own comparisons and from co-occurrence, not from a common experimental comparison. Risk of bias was appraised with a domain-based tool built for this review and not validated elsewhere; it was revised once after its initial version showed low agreement on a sample, before the final signalling-question version was applied across the corpus. The synthesis without meta-analysis counts the direction of effect against each report's own baseline; it does not measure effect size, and selective reporting inflates absolute proportions, so only differences between families are interpreted. Families were defined after extraction by rules on extracted fields, and the smallest, fastest-emerging families have wide intervals; the emergence index uses publication dates, which are approximate for records dated by year only.

\section{Conclusion}
\label{sec:conclusion}

Across 2,289 studies, AI routing in sparse MoE layers, model cascades, and multi-agent workflows consistently follows a shared inspect-then-allocate structure, even though input signals, resource constraints, and failure patterns diverge sharply across scales. Discrete hard choices remain the default everywhere ($70\%$, $83\%$, and $91\%$), while five mitigation families (router regularization, budget controls, calibration and verification, systems optimization, and modular design) appear across all three levels and represent promising candidates for cross-scale transfer.

At the same time, empirical claims require careful interpretation. Because individual papers compare against disparate self-selected baselines, family-level advantages carry low certainty and weaken among peer-reviewed publications. Research volume is shifting fastest toward multi-agent orchestration, yet that is precisely where routing mechanisms are least often ablated and risk of bias reaches its peak ($64\%$ among LLM orchestrators). Moving forward, three practical improvements would strengthen the field:
\begin{enumerate}
    \item \emph{Latency and systems-grounded measurement:} Token-level MoE studies should measure wall-clock latency and communication overhead alongside theoretical FLOPs, while query and agent studies should report end-to-end latency together with API cost.
    \item \emph{Direct routing baselines:} Multi-agent frameworks and query routers should evaluate delegation choices directly through router ablations, shared benchmarks such as RouterBench, and consistent reference baselines.
    \item \emph{Reporting consistency:} Routing studies should routinely include multi-seed variance and benchmark contamination checks so reported gains rest on reproducible comparisons.
\end{enumerate}


\appendix
\section{Protocol Reliability, Screening Marginals, and Amendment History}
\label{sec:appendix_protocol}

This appendix provides granular quality-assurance logs, agreement marginals, and protocol adaptations supplementing the methodological framework described in Section~\ref{sec:methods}.

\subsection{Screening Inter-Rater Reliability and Adjudication Criteria}
During title and abstract screening across all $N = 13{,}884$ deduplicated records, two review authors worked independently under dual-blind conditions. The resulting $2 \times 2$ inter-rater contingency matrix comprises $a = 6{,}896$ concordant inclusions, $d = 5{,}948$ concordant exclusions, $b = 826$ reviewer-1-only inclusions, and $c = 214$ reviewer-2-only inclusions (overall raw agreement 92.51\%), yielding Cohen's $\kappa = 0.850$ (95\% CI 0.841--0.859; PABAK 0.850). All 1,040 discordant records were resolved by consensus discussion with a third review author.

For full-text eligibility edge cases under criterion IN2 (\emph{routing evaluated as routing}), initial double-screening revealed boundary disagreements on 28 pilot reports. The author team established three explicit adjudication consensus rules:
\begin{enumerate}
    \item \emph{Downstream application boundary:} General multi-agent workflows or specialized pipelines for a single problem (e.g., medical report summarization, legal drafting) without comparative routing evaluation or algorithmic router characterization fail IN2.
    \item \emph{Routing mechanism vs.\ static scheduling:} Fixed round-robin calling, unchanged heuristic orderings, or simple majority voting without state- or input-dependent dispatch do not constitute dynamic routing.
    \item \emph{Theoretical analyses:} Rigorous theoretical analyses, mathematical formalizations, and bounds on router allocation or capacity optimization satisfy IN2 even in the absence of benchmark experiments.
\end{enumerate}

\subsection{Extraction Provenance and Evidence Quotation Audit}
Across all 2,289 included reports, full-text data extraction captured $18{,}273$ evidence quotations corroborating extracted categorical fields and directional effect claims. A random audit sample of quotations verified against the primary PDF source texts demonstrated high provenance fidelity: 50.8\% exact verbatim matches, 22.0\% near-verbatim matches (minor whitespace or punctuation variations), 25.6\% faithful semantic paraphrases, 0.4\% author-clarified passages, and 1.2\% contextual matches, achieving 100.0\% provenance accounting (Supplementary S3b).

\subsection{Protocol Amendment Chronology}
Table~\ref{tab:app_amendments} documents the complete record of protocol refinements executed between initial scoping and final synthesis (PRISMA 2020 item 24c).

\begin{table}[!htb]
\centering
\caption{Chronological record of protocol refinements and adaptations.}
\label{tab:app_amendments}
\tablesetup
\begin{tabularx}{\textwidth}{@{}>{\raggedright\arraybackslash}p{0.08\textwidth} >{\raggedright\arraybackslash}p{0.25\textwidth} >{\raggedright\arraybackslash}X@{}}
\toprule
\textbf{Ref.} & \textbf{Stage} & \textbf{Description and Methodological Rationale} \\
\midrule
A1 & Screening & IN2 refined before data extraction to clarify boundaries between core routing algorithms and peripheral downstream applications; operationalized with the ``two to exclude'' dual-consensus rule. \\
A2 & Extraction & Dropped planned extraction subset of highly cited papers; all 2,289 included reports were extracted from full text. The 386 most-cited reports were retained as the primary seed set for backward citation searching. \\
A3 & Citation search & Extended citation chasing from the 386-paper seed to all included reports, incorporating OpenAlex alongside Semantic Scholar to ensure indexing saturation. \\
A4 & Full text & Introduced mandatory dual-reviewer confirmation for full-text exclusions and established exclusion code EX9 for withdrawn/retracted preprints ($n=14$). \\
A5 & Corpus re-check & Following adjudication of the 28 borderline pilot edge cases, established formal consensus rules and executed a corpus-wide IN2 re-assessment, excluding 430 non-routing application studies. \\
A6 & Quality appraisal & Upgraded initial 5-domain reproducibility checklist to a tailored 6-domain risk-of-bias instrument (RoB-Route) with explicit signalling questions addressing ML benchmark comparability and leakage. \\
A7 & Synthesis & Incorporated systematic reporting-bias appraisals, sensitivity scenarios, and adapted GRADE certainty assessments alongside the synthesis without meta-analysis (SWiM). \\
A8 & Family synthesis & Formulated the 20 approach families, emergence metrics, and directional vote-counting post-extraction as an exploratory, hypothesis-generating synthesis. \\
\bottomrule
\end{tabularx}
\end{table}

\FloatBarrier

\section{Supplementary Sensitivity and Subgroup Analyses}
\label{sec:appendix_sensitivity}

This appendix provides detailed numerical tabulations of the sensitivity analyses and secondary position maps supporting Section~\ref{sec:results_families} and Section~\ref{sec:results_sensitivity}.

\subsection{Sensitivity Scenarios Across Methodological Subgroups}
Table~\ref{tab:app_sensitivity} presents odds ratios and cross-family correlations across five pre-specified sensitivity scenarios.

\begin{table}[!htb]
\centering
\caption{Sensitivity analyses of the approach-family synthesis across five methodological scenarios (Appendix~\ref{sec:appendix_sensitivity}). Odds ratios (95\% CI) for reporting an improvement, adjusted for year, against the reference family of the primary analysis (T5 learned top-$k$ gates; Q1 trained classifiers; A2 workflow or topology search). $\rho$: Spearman correlation across families between the shares reporting better quality and lower compute.}
\label{tab:app_sensitivity}
\tablesetup
\begin{tabularx}{\textwidth}{@{}>{\raggedright\arraybackslash}X r r r r r r@{}}
\toprule
\textbf{Scenario} & \textbf{Reports} & \textbf{$\rho$} & \textbf{T2 quality} & \textbf{T1 compute} & \textbf{Q5 compute} & \textbf{A1 compute} \\
\midrule
Primary & 2,289 & -0.54 & 2.92 (1.55--5.48) & 3.74 (1.90--7.36) & 0.44 (0.24--0.83) & 0.44 (0.20--0.94) \\
Named venue only & 1,058 & -0.13 & 5.85 (1.74--19.63) & 4.40 (1.46--13.27) & 0.52 (0.24--1.15) & 0.18 (0.04--0.95) \\
Dated to the month only & 2,020 & -0.50 & 3.00 (1.51--5.96) & 4.14 (1.97--8.70) & 0.38 (0.19--0.76) & 0.48 (0.22--1.08) \\
Full-text basis only & 2,286 & -0.53 & 2.92 (1.55--5.48) & 3.74 (1.90--7.36) & 0.44 (0.24--0.83) & 0.44 (0.20--0.94) \\
Excluding high risk of bias & 1,450 & -0.52 & 4.30 (1.51--12.25) & 4.24 (1.92--9.36) & 0.37 (0.17--0.81) & 0.57 (0.20--1.66) \\
\bottomrule
\end{tabularx}
\end{table}

\subsection{Position Map: Emergence vs.\ Lower Compute Benefit}
\begin{figure}[!htb]
\centering
\includegraphics[width=0.90\textwidth]{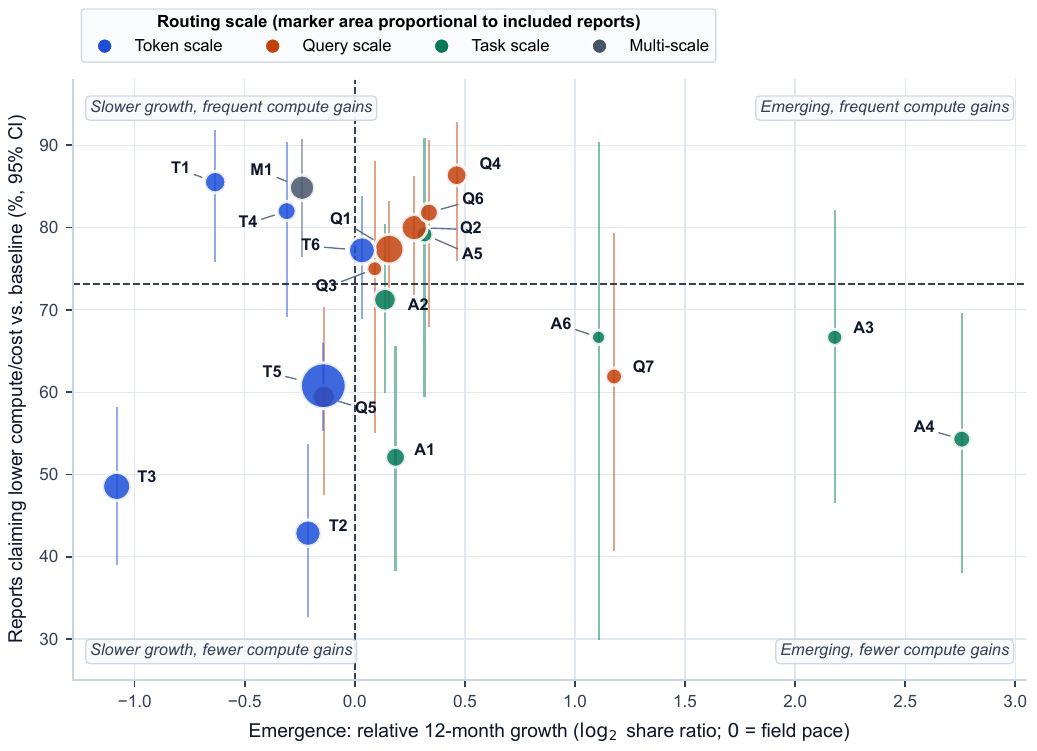}
\caption{Position map: emergence against the proportion of reports with lower compute or cost than their own baseline (95\% Wilson intervals). The dashed horizontal line is the median across families (Appendix~\ref{sec:appendix_sensitivity}).}
\label{fig:app_posmap2}
\end{figure}

Figure~\ref{fig:app_posmap2} displays approach-family emergence against the proportion of reports claiming lower compute or cost relative to their primary baseline.

\FloatBarrier

\section*{CRediT authorship contribution statement}

\textbf{Dat Tat Mai:} Conceptualization, Methodology, Software, Data curation, Formal analysis, Investigation, Validation, Writing -- original draft, Writing -- review and editing.

\textbf{Hung Tran Quy:} Methodology, Investigation, Validation, Writing -- review and editing.

\textbf{Thai Pham:} Conceptualization, Methodology, Investigation, Validation, Writing -- original draft, Writing -- review and editing.

\textbf{James Jin Kang:} Conceptualization, Supervision, Project administration, Resources, Writing -- review and editing.

\bibliographystyle{IEEEtran}
\renewcommand*{\bibfont}{\small}
\setlength{\bibsep}{3pt}
\bibliography{references}

\end{document}